\documentclass[10pt,twocolumn,letterpaper]{article}

\usepackage{iccv}
\usepackage{times}
\usepackage{epsfig}
\usepackage{graphicx}
\usepackage{amsmath}
\usepackage{amssymb}
\usepackage{array}
\usepackage{diagbox}
\usepackage{rotating}
\usepackage{multirow}
\usepackage{caption}
\usepackage{color}
\usepackage{booktabs}
\usepackage[unicode=true,
 bookmarks=false,
 breaklinks=true,pdfborder={0 0 1},backref=section,colorlinks=false]
 {hyperref}
\hypersetup{
 pagebackref=true,colorlinks}
 \usepackage[table,xcdraw]{xcolor}
 \usepackage[accsupp]{axessibility}  

\usepackage[breaklinks=true,bookmarks=false]{hyperref}

\iccvfinalcopy 

\def\iccvPaperID{3305} 

\ificcvfinal\fi

\begin{document}
\title{Learning a Sketch Tensor Space for Image Inpainting of Man-made Scenes} 
\author{Chenjie Cao,  Yanwei Fu\\
School of Data Science, Fudan University\\
{\tt\small \{20110980001,yanweifu\}@fudan.edu.cn}
}

\twocolumn[{%
\renewcommand\twocolumn[1][]{#1}%
\maketitle
\begin{center}
\centering
\vspace{-0.25in}
\includegraphics[width=0.85\linewidth]{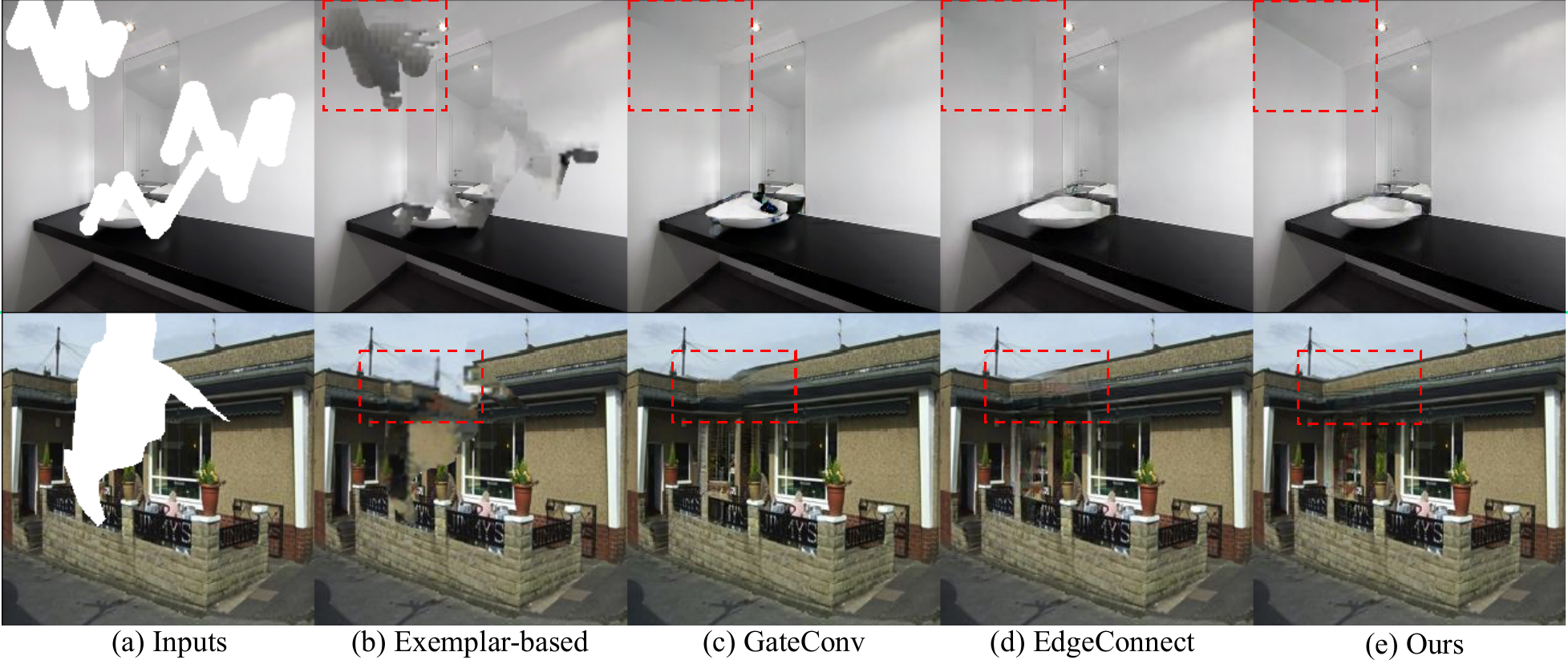}
\vspace{-0.1in}
\captionof{figure}{Results of (b) Exemplar-based inpainting~\protect\cite{criminisi2003object}, (c) GateConv~\protect\cite{yu2019free}, (d) EdgeConnect~\protect\cite{Nazeri_2019_ICCV}, and  (e) our method, with the inpainted areas highlighted by red dot box. Our method makes better inpainting without artifacts.}
\label{teaser} 
\end{center}
}]

\maketitle
\ificcvfinal\thispagestyle{empty}\fi

\begin{abstract} 
\vspace{-0.1in} 
This paper studies the task of inpainting man-made scenes. It is very challenging due to the difficulty in preserving the visual patterns of images, such as edges, lines, and junctions. Especially, most previous works are failed to restore the object/building structures for images of man-made scenes. To this end, this paper proposes learning a Sketch Tensor (ST) space for inpainting man-made scenes. Such a space is learned to restore the edges, lines, and junctions in images, and thus makes reliable predictions of the holistic image structures. To facilitate the structure refinement, we propose a Multi-scale Sketch Tensor inpainting (MST) network, with a novel encoder-decoder structure. The encoder extracts lines and edges from the input images to project them into an ST space. From this space, the decoder is learned to restore the input images. Extensive experiments validate the efficacy of our model. Furthermore, our model can also achieve competitive performance in inpainting general nature images over the competitors. 
\end{abstract} 
\vspace{-0.15in}

\section{Introduction}

As a long-standing problem, image inpainting has been studied to address
the problem of filling in the missing parts of the images being semantically
consistent and visually realistic with plausible results. Thus, image
inpainting is useful to many real-world applications, \textit{e.g.}
image restoration, image editing, and object removal~\cite{elharrouss2019image}.

Intrinsically as an {inverse problem}, the inpainting is challenging
in both restoring the missed global structure (\textit{semantically
consistent}), and generating realistic regions locally coherent to
unmasked regions (\textit{visually consistent}). Especially, it is
hard to reconstruct the missed image regions from the complex man-made
scenes and structures, due to the difficulty in preserving the prominent
low-level visual patterns, such as edges, line segments, and junctions,
as shown in Fig.~\ref{teaser}. To this end, this paper particularly
focuses on learning to reconstruct these visual patterns for image
inpainting, and proposes a method of the best merits in repairing
the masked regions of man-made scene images, such as images with indoor
and outdoor buildings.

Both traditional approaches~\cite{criminisi2003object,guo2017patch,li2017localization}
and deep learning methods~\cite{liu2018image,zeng2019learning,Yi_2020_CVPR,Lahiri_2020_CVPR,yu2019free,li2020recurrent} had made great efforts on reconstructing the structures of images
in producing visually realistic results. However, these methods are
still challenged by producing structurally coherent results, especially in
the inpainting of man-made scenes. 
Typically, inpainting approaches may suffer from the following problems.
(1)~\textit{Missing critical structures.} Traditional synthesis-based
approaches are normally unable to model the critical structures as
in Fig.~\ref{teaser}(b). On the other hand, recent learning-based
inpainting methods utilize auxiliary information to support the inpainting,
\textit{e.g.}, edges~\cite{Nazeri_2019_ICCV,li2019progressive},
and segmentation~\cite{song2018spg,liao2020guidance}, predominantly inpainting local visual cues, rather than holistic structures of man-made scenes. For example, the results of EdgeConnect~\cite{Nazeri_2019_ICCV}
with canny edges~\cite{ding2001canny} in Fig.~\ref{teaser}(d)
suffer from broken and blurry line segments and lose connectivity
of building structures. (2)~\textit{Unreliable pattern transfer.}
The learning-based auxiliary detectors will transfer and magnify the
unreliable image priors or patterns to the masked image
regions, which causes degraded inpainting results~\cite{liao2020guidance}.
(3) \textit{Trading off performance and efficiency.} The auxiliary-based
inpainting methods usually consume more computation, due to additional
components or training stages~\cite{Nazeri_2019_ICCV,li2019progressive,yang2020learning}.
But these methods still have artifacts in junction regions as framed
with red dotted lines in Fig.~\ref{teaser}(c)(d). Therefore, the design of a more
effective network is expected to efficiently enhance the inpainting performance.

To address these issues, our key idea is to learn a Sketch Tensor
(ST) space by an encoder-decoder model. (1) The encoder learns to
infer both local and holistic critical structures of input images,
including canny edges and compositional lines. The image is encoded
as the binarized `sketch' style feature maps, dubbed as \emph{sketch
tensor}. The decoder takes the restored structure to fill in holes
of images. 
(2) For the first time, the idea of parsing wireframes~\cite{huang2018learning}
is re-purposed to facilitate inpainting by strengthening the holistic
structures of man-made scenes with more effective and flexible holistic
structures. We propose a Line Segment Masking (LSM) algorithm to effectively
train the wireframe parser, which alleviates unreliable structure guidance from
corrupted images and the heavy computation of auxiliary detectors during the training phase. Besides, LSM also leverage the separability of line segments to extend the proposed model to obtain better object removal results.
(3) Most importantly, we significantly
boost the training and inference process of previous inpainting architectures.
Thus, a series of efficient modules are proposed, which include partially
gated convolutions, efficient attention module, and Pyramid Decomposing
Separable (PDS) blocks. Critically, we present PDS blocks to help
better learn binary line and edge maps. Our proposed modules make
a good balance of model performance and training efficiency.

\begin{figure*}
\begin{centering}
\includegraphics[width=0.85\linewidth]{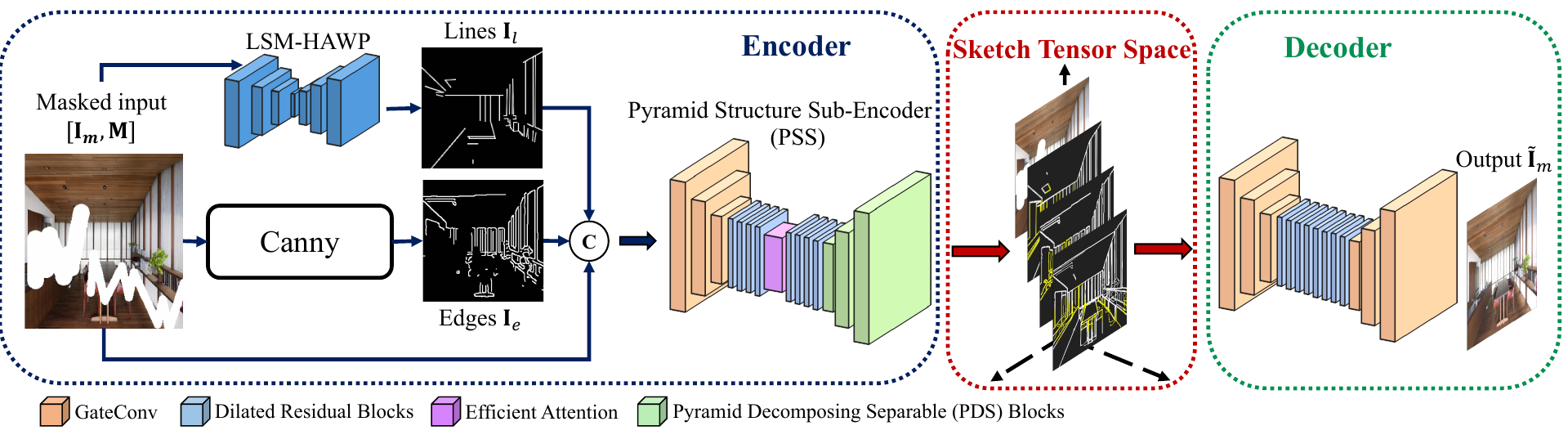} 
\par\end{centering}
\vspace{-0.15in}
 \caption{The overview of MST, which is consisted of encoder, sketch tensor
space, and decoder.}
\label{overview1} \vspace{-0.2in}
 
\end{figure*}

Formally, this paper proposes a novel Multi-scale Sketch Tensor inpainting
(MST) network with an encoder-decoder structure. The encoder employs
LSM algorithm to train an hourglass wireframe parser~\cite{xue2020holistically}
and a canny detector to extract line and edge maps. These maps concatenated
with input images are projected into ST space by Pyramid Structure
Sub-encoder (PSS), which is sequentially constructed by 3 partially gated convolution
layers, 8 dilated residual block layers with an efficient attention module,
and 3 pyramid decomposing block layers as in Fig.~\ref{overview1}.
The image is encoded as a third-order sketch tensor in ST space, representing local and holistic structures. Finally, the decoder is stacked by two groups of 3 partially gated convolution layers for both ends with 8 residual block layers, which will re-project the sketch tensor into the restored image. 

We highlight several contributions here.
(1) We propose learning a novel sketch tensor space for inpainting tasks. Such a space is learned to to restore the critical missed structures and visual patterns, and makes reliable predictions of the holistic image structures. Essentially, the skech tensor has good interpretability of input images,  as empirically shown in experiments.
(2) For the first time, the wireframe parsing has been re-purposed to extract lines and junctions for inpainting. A novel line segment masking algorithm is proposed to facilitate training our inpainting wireframe parser. 
(3) We introduce the novel partially gated convolution
and efficient attention module to significantly improve model performance
without incurring additional expensive computational cost than competitors,
such as EdgeConnect. 
(4) A novel pyramid decomposing separable block is proposed
to address the issue of effectively learning sparse binary edge and
line maps. 
(5)  Extensive experiments on the dataset
of both man-made and natural scenes, including ShanghaiTech~\cite{huang2018learning},
Places2~\cite{zhou2017places}, and York Urban~\cite{denis2008efficient}.
show the efficacy of our MST-net, over the competitors. 

\section{Related work}

\noindent \textbf{Image Inpainting by Auxiliaries}. The auxiliary
information of semantics and structures have been utilized to help
inpainting tasks in traditional methods, such as lines,
structures~\cite{huang2014image,barnes2009patchmatch}, and approximation
images~\cite{hays2008scene}. Recently, deep learning based approaches
take auxiliary information as the important prior, such as canny edges~\cite{Nazeri_2019_ICCV,li2019progressive},
smoothed edges~\cite{ren2019structureflow,Liu2019MEDFE}, edge and
gradient information~\cite{yang2020learning}, and semantic segmentation~\cite{song2018spg,liao2020guidance}.
Auxiliary information has been also utilized in image editing~\cite{zhang2017real,sangkloy2017scribbler,jo2019sc,lee2020maskgan}.
Particularly, EdgeConnect~\cite{Nazeri_2019_ICCV} and GateConv~\cite{yu2019free}
both leverage edges to inpaint masked areas with specific structural
results. However, there is no holistic structure information for man-made
scenes that has been explicitly modeled and utilized as auxiliary
information in previous work. Such information is crucial to inpaint
the images of complex structures, such as buildings, or indoor furniture.
To this end, our method utilizes the separability and connectivity
of line segments to help improve the performance of inpainting. 

\noindent \textbf{Line Detection Approaches}. Line detection enjoys
immense value in many real-world problems. Therefore, it has been
widely researched and developed in computer vision. Many classic graphics
algorithms are proposed to extract line segments from raw images~\cite{hart1972use,von2008lsd,lu2015cannylines}.
However, these traditional methods suffer from intermittent and spurious
line segments. Moreover, the extracted lines lack positions of junctions
which causes poor connectivity. Huang \etal~\cite{huang2018learning}
propose a deep learning based wireframe parsing to improve the line
detection task, which uses DNNs to predict heatmaps of junctions and
line segments as the wireframe representation. LCNN proposed in~\cite{zhou2019end}
leverages heuristic sampling strategies and a line verification network
to improve the performance of wireframe parsing. Furthermore, Xue
\etal~\cite{xue2020holistically} utilize the holistic attraction
field to enhance both efficiency and accuracy for the wireframe parser.
Although wireframe parsing based models enjoy distinctive strengths,
to the best of our knowledge, no previous work has considiered  utilizing wireframe parser to help preserve line structures for downstreaming  inpainting task. Critically,
for the first time, our method first repurposes the wireframe
parser to facilitate image inpainting, which leverages connectivity and
structural information of wireframes to achieve better inpainting
performance.

\section{Multi-scale Sketch Tensor Inpainting}

\begin{figure}
\begin{centering}
\includegraphics[width=0.9\linewidth]{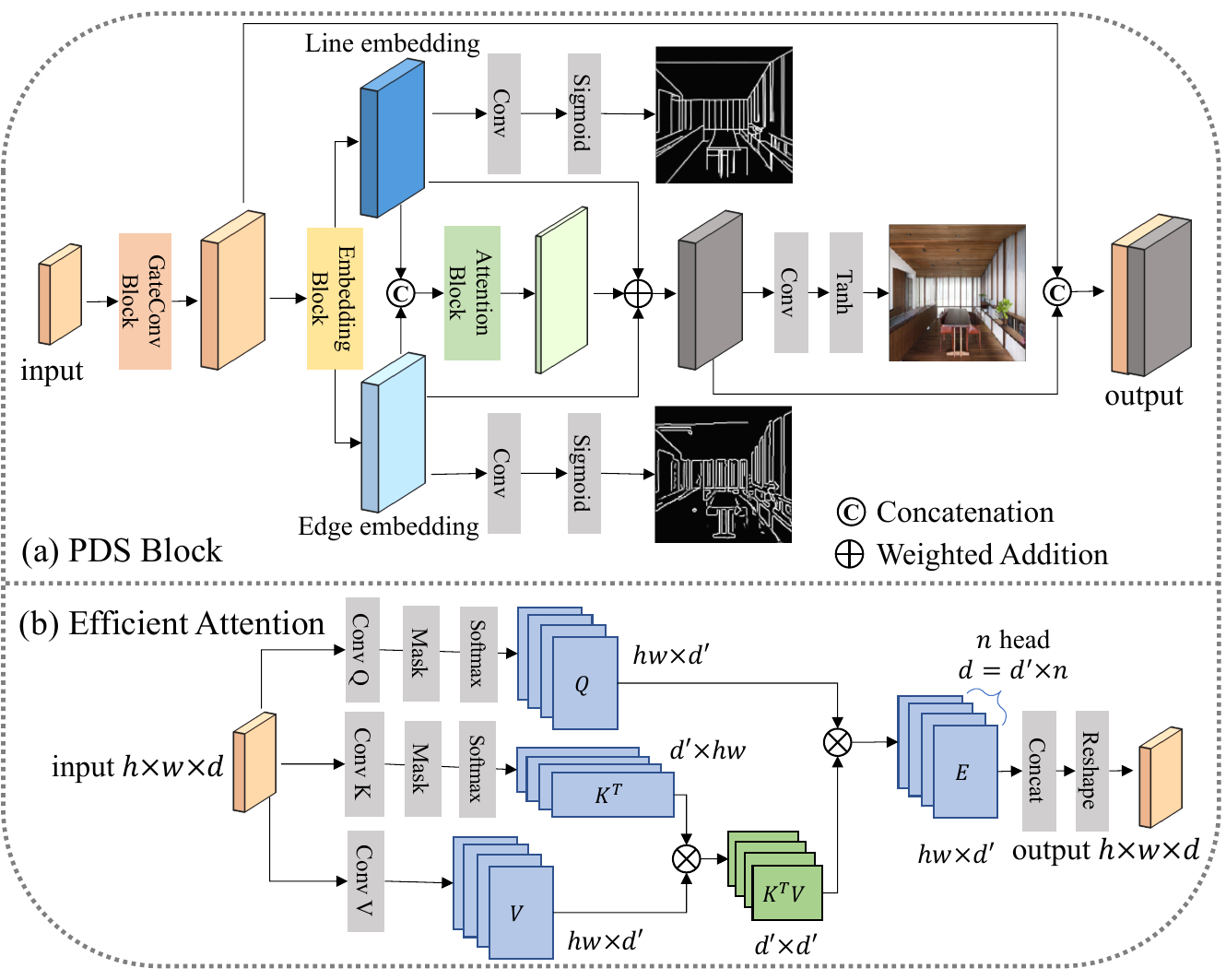} 
\par\end{centering}
\vspace{-0.1in}
 \caption{The illustration of (a): Pyramid Decomposing Separable (PDS) block,
(b): Efficient Attention block.}
\label{overview2} 
\vspace{-0.2in}
\end{figure}

\noindent \textbf{Overview.} The MST network is shown in Fig.~\ref{overview1}.
Given the input masked image $\mathbf{I}_{m}\in\mathbb{R}^{h\times w\times3}$
and corresponding binary mask $\mathbf{M}$, MST has three key components,
\textit{i.e.}, encoder $\mathbf{\Phi}:\left[\mathbf{I}_{m},\mathbf{M}\right]\rightarrow\mathcal{S}$, decoder $\mathbf{\Psi}:\mathcal{S}\rightarrow\tilde{\mathbf{I}}_{m}$, and Sketch Tensor (ST) space of a third-order tensor denoted by 
$\mathcal{S}\in\mathbb{R}^{h\times w\times 3}$.  
Particularly, the encoder firstly employs the improved wireframe parser LSM-HAWP and canny detector~\cite{ding2001canny} to extract line $\mathbf{I}_{l}$ and edge maps $\mathbf{I}_{e}$; then the concatenated image and maps
$\left[\mathbf{I}_{l};\mathbf{I}_{e};\mathbf{I}_{m};\mathbf{M}\right]$ is processed
by Pyramid Structure Sub-Encoder (PSS) to produce the 
ST space $\mathcal{S}$. The decoder predicts  inpainted image
$\tilde{\mathbf{I}}_{m}$, closer to ground-truth image $\mathbf{I}$.

In this section, Sec.~\ref{sec:LSM} will introduce the LSM algorithm. Details about the PSS and ST space
are specified in Sec.~\ref{sec:PSS}. Finally, the decoder will be discussed in Sec.~\ref{sec:IID}.

\subsection{Line Segment Masking Algorithm\label{sec:LSM}}

\begin{figure}
\begin{centering}
\includegraphics[width=0.85\linewidth]{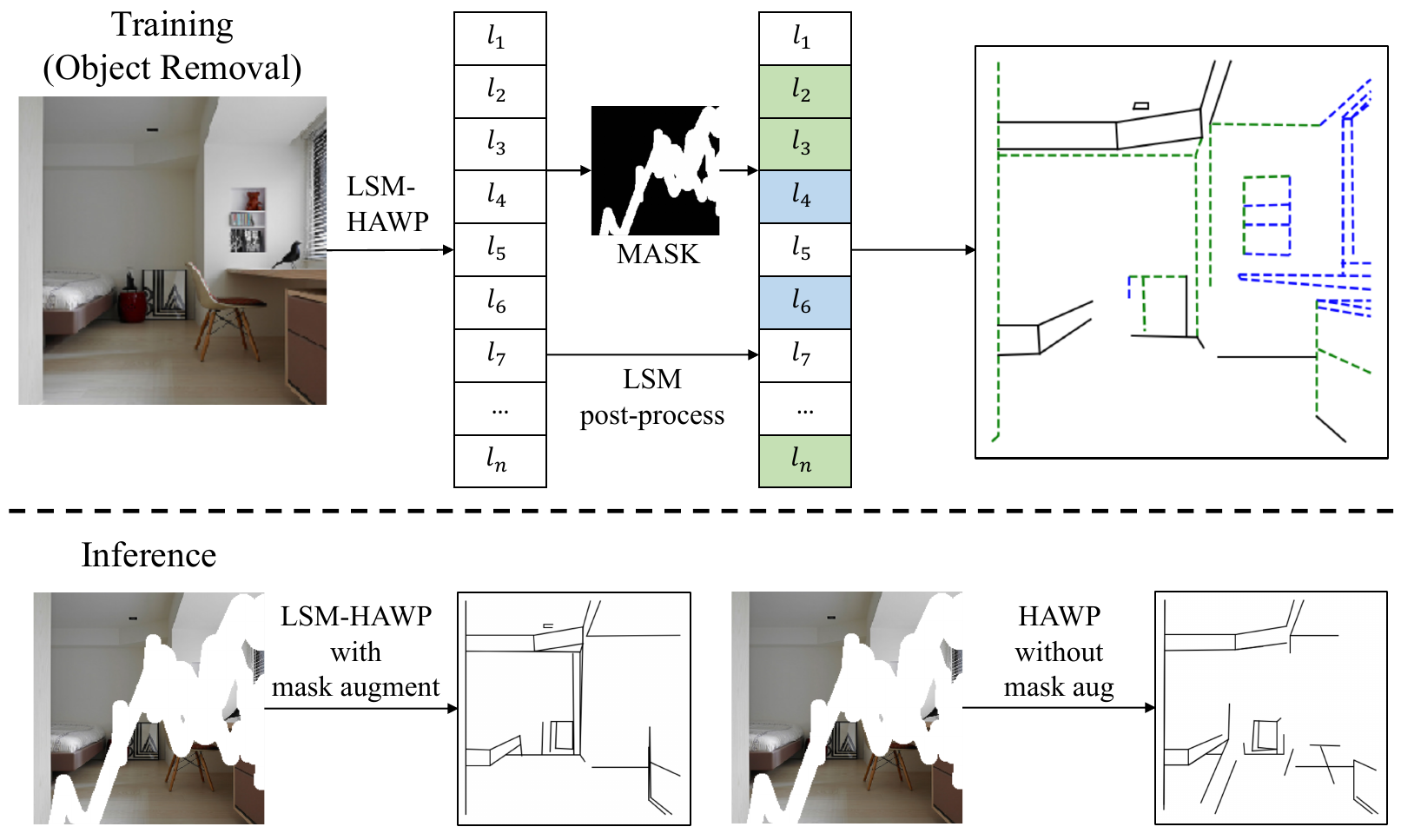} 
\par\end{centering}
\vspace{-0.1in}
 \caption{LSM Illustration. In training, lines segments are denoted
by Eq.~(\ref{2}). Black solid lines, blue and green dotted lines indicate the retained,
masked, and masked by probably $m$ lines. Only in
 inpainting training process or object removal task, the ground-truth
images are known in advance.}
\vspace{-0.2in}
 \label{f2} 
\end{figure}

The wireframe parser HAWP~\cite{xue2020holistically}
is adopted to extract lines from images. Specifically, it extracts the junction set $\mathbf{J}=\left\{ \mathbf{p}=\left(x,y\right)\in\mathbb{R}^{2}\right\} $,
and line set $\mathbf{L}=\left\{ \mathbf{l}=(\mathbf{p}_{a},\mathbf{p}_{b})=(x_{a},y_{a},x_{b},y_{b})\right\} $
paired by junctions in $\mathbf{J}$. Unfortunately, if lines are
corrupted by the mask $\mathbf{M}$, the results of naive HAWP~\cite{xue2020holistically}
will be largely degraded as shown in the inference stage of Fig.~\ref{f2}
with broken structures. 

To this end, we propose an LSM algorithm to infer missed line segments with the flexible wireframe parsing, which is composed of two parts. 
1) Learning an LSM-HAWP network by retraining
HAWP with the irregular~\cite{yu2019free} and object segmentation~\cite{zeng2020high}
masks. This not only improves the stability for the corrupted images, but also achieves superior results in trivial wireframe parsing tasks (Sec.~\ref{sec:Masked Wireframe Detection}). 
2) Introducing an indicator function to denote the masking probability
of each line segment $\mathbf{l}$ according to the mask $\mathbf{M}$ as the post-processing for LSM-HAWP,
\vspace{-0.05in}
\begin{equation}
\vspace{-0.05in}\mathbb{I}(\mathbf{l})=\begin{cases}
1~ & if\ \mathbf{p}_{a}\in\mathbf{M}~and~\mathbf{p}_{b}\in\mathbf{M}\\
0~ & if\ \mathbf{p}_{a}\notin\mathbf{M}~and~\mathbf{p}_{b}\notin\mathbf{M}\\
m~ & ~otherwise
\end{cases},\label{2}
\end{equation}
where $m~(0\leq m\leq1)$ is a hyper-parameter between 0 and 1, and
junction $\mathbf{p}\in M,\mathbf{p}\notin M$ means that $\mathbf{p}$ is masked or not. Fig.~\ref{f2} illustrates the LSM post-process in training and inference stages.
In training, we set $m=0.5$ to train the first half epoches, and $m=1$ for
training in the rest epoches. In the object removal task, we use the
unmasked image with $m=0$ to retain necessary structures. 

Thus as one practical strategy of speedup the learning process, we extract all wireframes beforehand and filtering them by the post-process of LSM, rather than an end-to-end training with LSM-HAWP, which dramatically improves the training efficiency. Note that in the testing stage for inpainting, corrupted images are used as the input to make our results fair comparable to the competitors.

\subsection{Pyramid Structure Sub-Encoder\label{sec:PSS}}

The LSM-HAWP and canny detector extract the line and edge maps $\mathbf{I}_{l},\mathbf{I}_{e}\in\mathbb{R}^{h\times w\times1}$ respectively.
Essentially, $\mathbf{I}_{e}$ is binary map, and $\mathbf{I}_{l}$ is got from connecting junction pairs from LSM-HAWP with anti-aliased lines.
The input to PSS is the concatenation of the masked image and structure maps $\left[\mathbf{I}_{l};\mathbf{I}_{e};\mathbf{I}_{m}\right]$. As shown in Fig. \ref{overview1}, the PSS is composed of partially gated convolutions, dilated~\cite{yu2015multi} residual blocks, efficient attention, and pyramid decomposing separable blocks, which will be explained next. For the detailed structures, please refer to our supplementary.

\noindent \textbf{Partially Gated Convolutions.} We adopt the Gated Convolution
(GC) layers to process the masked input features, as it works
well for the irregular mask inpainting tasks in ~\cite{yu2019free}.
Unfortunately, GC demands much more trainable parameters than vanilla convolutions. 
To this end, as shown in Fig.~\ref{overview1}
and Fig.~\ref{overview2}(a), we propose a partially GC strategy
of only utilizing three GC layers for the input and output features 
in both encoder and decoder models. Essentially, this
is motivated by our finding that the outputs of GC mostly devoting
to filtering features of masked and unmasked regions only in the encoder
layers of the coarse network and the decoder layers of the refinement
network in~\cite{yu2019free}. In contrast, we do not observe significantly
performance improvement of using GC in the middle layers of backbones.
Thus, we maintain vanilla convolutions (i.e., residual blocks) in
the middle layers, to save parameters and improve the performance
as empirically validated in Tab.~\ref{t4}.

\noindent \textbf{Efficient Attention Block.} Intuitively, attention
is important to learn patterns crossing spatial locations in image inpainting~\cite{yu2018generative,liu2019coherent,li2020recurrent,yang2020learning}. 
However, attention modules are expensive to be computed, and non-trivial
to be parallelized in image inpainting~\cite{yu2018generative}. To this end, we leverage
Efficient Attention (EA) module among middle
blocks of Fig. \ref{overview1} and detailed in Fig.~\ref{overview2}(b). Particularly,
for the input feature $\mathbf{X}\in\mathbb{R}^{h\times w\times d}$,
and mask $\mathbf{M}\in\mathbb{R}^{h\times w\times1}$, we first reshape
them to $\mathbb{R}^{hw\times d}$ and $\mathbb{R}^{hw\times1}$ as
the vector data respectively. Then, the process of efficient attention
can be written as 
\vspace{-0.07in} 
\begin{equation}
\vspace{-0.07in}\begin{split}\mathbf{Q} & ={\mathrm{softmax}}_{row}(\mathbf{W}_{q}\mathbf{X}+(\mathbf{M}\cdot-\infty))\\
\mathbf{K} & ={\mathrm{softmax}}_{col}(\mathbf{W}_{k}\mathbf{X}+(\mathbf{M}\cdot-\infty))\\
\mathbf{V} & =\mathbf{W}_{v}\mathbf{X},\;\;\;\;\;\;\;\mathbf{E}=\mathbf{Q}(\mathbf{K}^{T}\mathbf{V}),
\end{split}
\label{4}
\end{equation}
where $\mathbf{W}_{q,k,v}$ indicates different learned parameter matrices;
$+(\mathbf{M}\cdot-\infty)$ means masking the corrupted inputs before
the softmax operation. Critically, $\mathrm{softmax}_{row}$ and $\mathrm{softmax}_{col}$
means the softmax operations on the row and column individually. Then,
we achieve the output $\mathbf{E}\in\mathbb{R}^{hw\times d}$, which
will be reshaped back to $\mathbb{R}^{h\times w\times d}$. 
Note that Eq.~(\ref{4}) is an approximation to vanilla attention operation
as in ~\cite{shen2018efficient}. Typically, this strategy should
in principle, reduce significantly computational cost. Since $\mathbf{E}$
is computed by $\mathbf{K}^{T}\mathbf{V}\in\mathbb{R}^{d\times d}$,
rather than standard $\mathbf{Q}\mathbf{K}^{T}\in\mathbb{R}^{hw\times hw}$.
In practice, the dimension $d$ is much smaller than $hw$ in
computer vision tasks. Furthermore, as in Fig.~\ref{overview2}(b),
we introduce the multi-head attention~\cite{vaswani2017attention}
to further reduce the dimension from $d$ to $d'=d/n_{head}$.
Thus, $\mathbf{K}^{T}\mathbf{V}$ could aggregate the query $\mathbf{Q}$
in feature level and obtain the global context $\mathbf{E}$.
Note that EA module is inspired but different from ~\cite{shen2018efficient},
as the attention scores of corrupted regions are masked to aggregate
features from uncorrupted regions.

\noindent \textbf{Pyramid Decomposing Separable (PDS) Block.\label{sec:PDS}} We have found
that GAN-based edge inpainting~\cite{wang2018high} suffers from generating meaningless edges or unreasonable blanks for masked areas as shown in Fig.~\ref{f6}, due to the nature of sparsity in binary edge maps. 
Thus, we propose a novel PDS block in PSS as in Fig.~\ref{overview2}(a). PDS leverages
dilated lower-scale structures to make up the sparse problem for ST space.
Particularly, assume the image feature $\mathbf{X}\in\mathbb{R}^{h\times w\times d}$,
PDS firstly decouples it as
\vspace{-0.05in}
\begin{equation}
\vspace{-0.05in}
\begin{split}\mathbf{E}_{le} & =f_{eb}(\mathrm{GateConv}(\mathbf{X})),\;\{\mathbf{E}_{l},\mathbf{E}_{e}\}=\mathrm{split}(\mathbf{E}_{le}),\\
\negthickspace\mathbf{O}_{e} & =\sigma(\mathrm{Conv}(\mathbf{E}_{e})),\;\mathbf{O}_{l}=\sigma(\mathrm{Conv}(\mathbf{E}_{l})),
\end{split}
\label{7}
\end{equation}
where $\sigma$ is sigmoid activation. We project the feature
into two separated embedding spaces called line embedding $\mathbf{E}_{l}$
and edge embedding $\mathbf{E}_{e}$ with the function $f_{eb}$, which is composed with {\small{}$\mathrm{Conv2D\rightarrow IN\rightarrow ReLU}$}.
$\mathrm{split}$ is the operation of splitting the features into
two respective tensors with equal channels. 
Then, $\mathbf{E}_{l}$ and $\mathbf{E}_{e}$ are utilized to learn
to roughly reconstruct the image $\mathbf{O}_{im}$ without corruption
as
\vspace{-0.05in}
\begin{equation}
\vspace{-0.05in}\begin{split}\mathbf{A} & =f_{ab}(\mathbf{E}_{le})\in\mathbb{R}^{h\times w\times1}\\
\mathbf{E}_{le}' & =\mathbf{E}_{l}\odot(1-\mathbf{A})+\mathbf{E}_{e}\odot\mathbf{A}\\
\mathbf{O}_{im} & =\mathrm{tanh}(\mathrm{Conv}(\mathbf{E}_{le}')),
\end{split}
\label{8}
\end{equation}
where $f_{ab}$ is {\small{}$\mathrm{\mathrm{Conv2D}\rightarrow IN\rightarrow ReLU\rightarrow Conv2D\rightarrow Sigmoid}$}. $\odot$ denotes the element-wise product. The prediction of coarse
$\mathbf{O}_{im}$ gives a stronger constraint to the conflict of
$\mathbf{E}_{l}$ and $\mathbf{E}_{e}$, and we expect the attention
map $\mathbf{A}$ can be learned adaptively according to a challenging
target. Furthermore, a multi-scale strategy is introduced in the PDS.
For the given feature $\mathbf{X}^{(1)}\in\mathbb{R}^{64\times64\times d}$
from dilated residual blocks of PSS, PDS predicts lines, edges, and coarse
images with different resolutions as follow 
\vspace{-0.05in}
\begin{equation}
\vspace{-0.05in}\mathbf{X}^{(i+1)},\mathbf{O}_{im}^{(i)},\mathbf{O}_{l}^{(i)},\mathbf{O}_{e}^{(i)}=\mathrm{PDS}^{(i)}(\mathbf{X}^{(i)}),
\label{PDS_outputs}
\end{equation}
where $i=1,2,3$ indicate that the output maps with {\small{}64$\times$64,
128$\times$128, }and {\small{}256$\times$256} resolutions respectively.
Various scales of images can be in favor of reconstructing different
types of structure information.

\noindent \textbf{Loss Function of PSS.} We minimize the objectives of PPS with two spectral norm~\cite{miyato2018spectral} based SN-PathGAN~\cite{yu2019free} discriminators $D_l$ and $D_e$ for lines and edges respectively. And the $\mathcal{L}_{D}^{enc}$ and $\mathcal{L}_{G}^{enc}$ are indicated as
\vspace{-0.05in}
\begin{equation}
\vspace{-0.1in}
\begin{split}\mathcal{L}_{D}^{enc}= & \mathcal{L}_{D_{l}}+\mathcal{L}_{D_{e}},\\
\mathcal{L}_{G}^{enc}= & \lambda_{a}\mathcal{L}_{adv}^{enc}+\lambda_{f}\mathcal{L}_{fm}+\sum_{i=1}^{3}\lVert\mathbf{O}_{im}^{(i)}-\mathbf{I}^{(i)}\rVert_{1},
\end{split}
\label{13}
\end{equation}
where $\mathbf{O}_{im}^{(i)}$, $\mathbf{I}^{(i)}$ are the multi-scale
outputs (Eq.~\ref{PDS_outputs}) and ground-truth images varying image size {\small $64\times 64$ }
to {\small $256\times 256$ } respectively. $\mathcal{L}_{fm}$ is the feature matching loss
as in \cite{Nazeri_2019_ICCV} to restrict the $l_1$ loss between discriminator features from the concatenated coarse-to-fine real and fake sketches discussed below. And the adversarial loss can be further specified as
\vspace{-0.05in}
\begin{equation}
\vspace{-0.05in}
\begin{split}
\mathcal{L}&_{adv}^{enc}=-\mathbb{E}\left[\log D_{l}(\hat{\mathbf{O}}_{l})\right]-\mathbb{E}\left[\log D_{e}(\hat{\mathbf{O}}_{e})\right],\\
\mathcal{L}&_{D_{l}} =-\mathbb{E}\left[\log D_{l}(\hat{\mathbf{I}}_{l})\right]-\mathbb{E}\left[1-\log D_{l}(\hat{\mathbf{O}}_{l})\right],\\
\mathcal{L}&_{D_{e}} =-\mathbb{E}\left[\log D_{e}(\hat{\mathbf{I}}_{e})\right]-\mathbb{E}\left[1-\log D_{e}(\hat{\mathbf{O}}_{e})\right],
\end{split}
\label{9}
\end{equation}
where $\hat{\mathbf{O}}_{l}$, $\hat{\mathbf{O}}_{e}\in\mathbb{R}^{256\times256\times3}$
are got from the multi-scale PSS outputs of lines and edges, which are upsampled and concatenated with $\hat{\mathbf{O}}_{l}=[\mathrm{up}(\mathbf{O}_{l}^{(1)});\mathrm{up}(\mathbf{O}_{l}^{(2)});\mathbf{O}_{l}^{(3)}]$
and $\hat{\mathbf{O}}_{e}=[\mathrm{up}(\mathbf{O}_{e}^{(1)});\mathrm{up}(\mathbf{O}_{e}^{(2)});\mathbf{O}_{e}^{(3)}]$. To unify these outputs into the same scale, the nearest upsampling $\mathrm{up()}$ is utilized here.
Accordingly, $\hat{\mathbf{I}}_{l}$, and $\hat{\mathbf{I}}_{e}$ are the concatenation of multi-scale ground-truth edge and line maps, respectively. Note that, we firstly dilate the ground-truth edges and lines with the {\small{}$2\times2$} kernel, and then subsample them to produce the low resolution maps at the scale of {\small{}64$\times$64, 128$\times$128.}
The hyper-parameters are set as $\lambda_{a}=0.1$, $\lambda_{f}=10$ .

\subsection{Sketch Tensor (ST) Space}

The last outputs $\mathbf{O}_{l}^{(3)},\mathbf{O}_{e}^{(3)}\in\mathbb{R}^{256\times256\times1}$
of Eq.~\ref{PDS_outputs} is used to compose the ST space as 
\vspace{-0.05in}
\begin{equation}
\vspace{-0.05in}
\mathcal{S}=\left[\mathbf{O}_{l}^{(3)};\mathbf{O}_{e}^{(3)};\mathrm{clip}(\mathbf{O}_{l}^{(3)}+\mathbf{O}_{e}^{(3)})\right].\label{ST}
\end{equation}
Generally, lines represent holistic structures, while edges indicate
some local details. They provide priors of structures in different manners to the inpainting model. 
We also combine and clip them within 0 and 1 to
emphasize overlaps and make an intuitive expression of the whole structure.

\subsection{Decoder of MST Network}
\label{sec:IID} The structure of the decoder is
the same as PSS except that the decoder has no attention blocks and
PSD blocks. Because we find that the generator in~\cite{Nazeri_2019_ICCV}
is sufficient to generate fine results with reasonable structure information.
But we still add gated mechanism to the leading and trailing three
convolutions to improve the performance for the irregular mask. 
As the encoder has provided the ST space $\mathcal{S}\in\mathbb{R}^{h\times w\times3}$
in Eq.~\ref{ST}, we can get the inpainted image $\tilde{\mathbf{I}}_m$
with the masked input ${\mathbf{I}_m}$ and mask $\mathbf{M}$ as 
\vspace{-0.05in}
\begin{equation}
\vspace{-0.05in}
\tilde{\mathbf{I}}_m=\mathbf{\Psi}([\mathbf{I}_m;\mathbf{M};\mathcal{S}]).\label{14}
\end{equation}
We use SN-PathGAN~\cite{yu2019free} based discriminator $D_{im}$ for the decoder, and the objectives to be minimized are 
\vspace{-0.05in} 
\begin{equation}
\vspace{-0.05in} 
\begin{split}
\mathcal{L}_{D}^{dec}&=-\mathbb{E}\Big[\log D_{im}({\mathbf{I}})\Big]-\mathbb{E}\left[1-\log D_{im}(\tilde{\mathbf{I}}_m)\right],\\
\mathcal{L}_{G}^{dec}&=\mathcal{L}_{l_{1}}+\lambda_{a}\mathcal{L}_{adv}^{dec}+\lambda_{p}\mathcal{L}_{per}+\lambda_{s}\mathcal{L}_{style},
\label{20}
\end{split}
\end{equation}
\vspace{-0.05in} 
\begin{equation}
\vspace{-0.05in} 
\mathcal{L}_{adv}^{dec}=-\mathbb{E}\left[\log D_{im}(\tilde{\mathbf{I}}_m)\right],
\label{16}
\end{equation} 
where $\mathbf{I}$ is the origin undamanged image for training. We adopt the 
$l_{1}$ loss $\mathcal{L}_{l_{1}}$, VGG-19 based perceptual loss~\cite{johnson2016perceptual}
$\mathcal{L}_{per}$, and style-loss~\cite{gatys2016image} $\mathcal{L}_{style}$ to train our model, with the empirically setted hyperparameters $\lambda_{p}=0.1$, $\lambda_{s}=250$~\cite{Nazeri_2019_ICCV,Liu2019MEDFE}.
We propose using the balanced loss form~\cite{cui2019class} to implement reconstruction losses $\mathcal{L}_{l_{1}}$ and $\mathcal{L}_{per}$, normalizing by the weights of masked region $\mathbf{M}$ and  unmasked region $1-\mathbf{M}$.
The balanced loss form can settle the inpainting task with imbalanced irregular masks properly. 

Note that we learn the encoder and the decoder jointly in the forward pass. The gradients
of the decoder will not be propagated to update the encoder, which not only saves the memory but also maintains the interpretability of the sketch tensor $\mathcal{S}$. Generally, the total parameter number and calculation costs of our model are comparable to~\cite{Nazeri_2019_ICCV}.

\section{Experiments and Results}

\begin{figure*}
\begin{centering}
\includegraphics[width=0.85\linewidth]{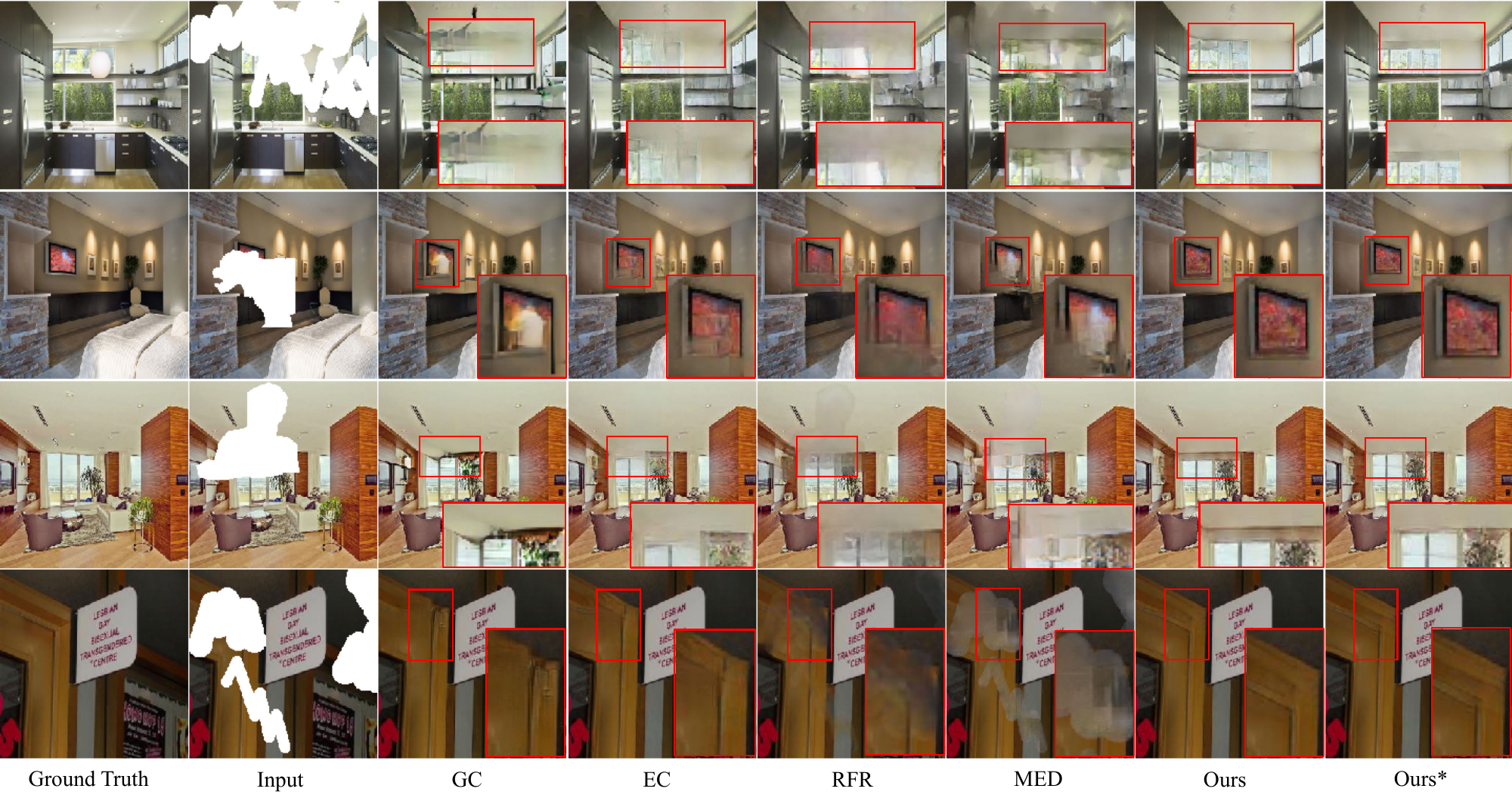} 
\par\end{centering}
\vspace{-0.1in}
 \caption{Qualitative results in ShanghaiTech and man-made Places2, where {*}
means that our method works in the LMS object removal mode with $m=0$.
Key parts are enlarged, and the complete generated pictures are shown
in the supplementary.}
\label{f3} \vspace{-0.15in}
\end{figure*}

\begin{table}
\begin{centering}
\renewcommand\tabcolsep{5.5pt}
\begin{tabular}{c}
{\small{}{}{}{}\hspace{-0.07in}} 
\begin{tabular}{c|c|cccccc}
\hline 
 &  & {\small{}{}{}GC}  & {\small{}{}{}EC}  & {\small{}{}{}RFR}  & {\small{}{}{}MED}  & {\small{}{}{}Ours}  & {\small{}{}{}Ours{*}}\tabularnewline
\hline 
\hline 
\multirow{3}{*}{\begin{turn}{90} {\small{}{}{}S.-T.} \end{turn}} & {\small{}{}{}P.$\uparrow$}  & {\small{}{}{}25.54}  & {\small{}{}{}26.51}  & {\small{}{}{}26.30}  & {\small{}{}{}26.14}  & \textbf{\small{}{}{}26.90}{\small{}{} }  & {\small{}{}{}27.20}\tabularnewline
\cline{2-8} \cline{3-8} \cline{4-8} \cline{5-8} \cline{6-8} \cline{7-8} \cline{8-8} 
 & {\small{}{}{}S.$\uparrow$}  & \textbf{\small{}{}{}0.893}{\small{}{} }  & {\small{}{}{}0.871}  & {\small{}{}{}0.863}  & {\small{}{}{}0.886}  & {\small{}{}{}0.876}  & {\small{}{}{}0.880}\tabularnewline
\cline{2-8} \cline{3-8} \cline{4-8} \cline{5-8} \cline{6-8} \cline{7-8} \cline{8-8} 
 & {\small{}{}{}F.$\downarrow$}  & {\small{}{}{}26.06}  & {\small{}{}{}24.84}  & {\small{}{}{}25.95}  & {\small{}{}{}30.02}  & \textbf{\small{}{}{}21.16}{\small{}{} }  & {\small{}{}{}20.05}\tabularnewline
\hline 
\multirow{3}{*}{\begin{turn}{90} {\small{}{}{}P2M} \end{turn}} & {\small{}{}{}P.$\uparrow$}  & {\small{}{}{}25.72}  & {\small{}{}{}26.71}  & {\small{}{}{}26.07}  & {\small{}{}{}23.84}  & \textbf{\small{}{}{}27.20}{\small{}{} }  & {\small{}{}{}27.47}\tabularnewline
\cline{2-8} \cline{3-8} \cline{4-8} \cline{5-8} \cline{6-8} \cline{7-8} \cline{8-8} 
 & {\small{}{}{}S.$\uparrow$}  & {\small{}{}{}0.892}  & {\small{}{}{}0.901}  & {\small{}{}{}0.890}  & {\small{}{}{}0.857}  & \textbf{\small{}{}{}0.907}{\small{}{} }  & {\small{}{}{}0.910}\tabularnewline
\cline{2-8} \cline{3-8} \cline{4-8} \cline{5-8} \cline{6-8} \cline{7-8} \cline{8-8} 
 & {\small{}{}{}F.$\downarrow$}  & {\small{}{}{}16.54}  & {\small{}{}{}14.75}  & {\small{}{}{}17.79}  & {\small{}{}{}26.95}  & \textbf{\small{}{}{}12.81}{\small{}{} }  & {\small{}{}{}12.13}\tabularnewline
\hline 
\multirow{3}{*}{\begin{turn}{90} {\small{}{}{}Y.-U.} \end{turn}} & {\small{}{}{}P.$\uparrow$}  & {\small{}{}{}25.92}  & {\small{}{}{}26.13}  & {\small{}{}{}25.64}  & {\small{}{}{}24.20}  & \textbf{\small{}{}{}26.29}{\small{}{} }  & {\small{}{}{}26.59}\tabularnewline
\cline{2-8} \cline{3-8} \cline{4-8} \cline{5-8} \cline{6-8} \cline{7-8} \cline{8-8} 
 & {\small{}{}{}S.$\uparrow$}  & {\small{}{}{}}\textbf{\small{}{}0.886}{\small{} } & {\small{}{}{}0.864}  & {\small{}{}{}0.852}  & {\small{}{}{}0.858}  & {\small{}{}{}0.869}  & {\small{}{}{}0.872}\tabularnewline
\cline{2-8} \cline{3-8} \cline{4-8} \cline{5-8} \cline{6-8} \cline{7-8} \cline{8-8} 
 & {\small{}{}{}F.$\downarrow$}  & {\small{}{}{}31.68}  & {\small{}{}{}32.06}  & {\small{}{}{}38.70}  & {\small{}{}{}51.71}  & \textbf{\small{}{}{}29.15}{\small{}{} }  & {\small{}{}{}26.53}\tabularnewline
\hline 
\multirow{3}{*}{\begin{turn}{90} {\small{}{}{}P2C} \end{turn}} & {\small{}{}{}P.$\uparrow$}  & {\small{}{}{}27.87}  & {\small{}{}{}28.35}  & --  & --  & \textbf{\small{}{}{}28.52}{\small{}{} }  & {\small{}{}{}28.65}\tabularnewline
\cline{2-8} \cline{3-8} \cline{4-8} \cline{5-8} \cline{6-8} \cline{7-8} \cline{8-8} 
 & {\small{}{}{}S.$\uparrow$}  & {\small{}{}{}0.923}  & {\small{}{}{}0.927}  & --  & --  & {\small{}{}{}}\textbf{\small{}{}0.928}{\small{} } & {\small{}{}{}0.929}\tabularnewline
\cline{2-8} \cline{3-8} \cline{4-8} \cline{5-8} \cline{6-8} \cline{7-8} \cline{8-8} 
 & {\small{}{}{}F.$\downarrow$}  & {\small{}{}{}15.05}  & {\small{}{}{}13.68}  & --  & --  & \textbf{\small{}{}{}11.97}{\small{}{} }  & {\small{}{}{}11.69}\tabularnewline
\hline 
\end{tabular}\tabularnewline
\end{tabular}
\par\end{centering}
\vspace{-0.1in}
 \caption{The PSNR (P.), SSIM (S.) and FID (F.) results on ShanghaiTech (S.-T.),
man-made Places2 (P2M), York Urban (Y.-U.), and comprehensive Places2
(P2C). $\uparrow$ means larger is better and $\downarrow$ means
lower is better. {*} indicates our method working in object removal
mode of LSM. Best results are bold except for object removal ones.}
\label{t2} \vspace{-0.2in}
\end{table}

\noindent \textbf{Datasets.} The proposed approach is evaluated on
three datasets: ShanghaiTech~\cite{huang2018learning}, Places2~\cite{zhou2017places},
and York Urban~\cite{denis2008efficient}. ShanghaiTech contains
5000 training images and 462 test images consisted of buildings and
indoor scenes with wireframe labels. The LSM-HAWP wireframe detection
model is trained on this dataset with the mask augmentation.
For Places2, we select 10 categories images according to the number
of line segments detected by HAWP as man-made scenes (P2M). Moreover,
we randomly select 10 Places2 classes with both natural and urban
scenes as comprehensive scenes (P2C)\footnote{Details about Places2 are illustrated in the supplementary.}.
For York Urban, it has 102 city street view images for testing models
trained with P2M.

\noindent \textbf{Implementation Details.} Our method is implemented
with PyTorch in {\small{}$256\times256$} image size. All models are trained
with Adam optimizer of $\beta_{1}=0$ and $\beta_{2}=0.9$, and the
initial learning rates are {\small{}$2e-4$} and {\small$2e-5$} for generators and
discriminators respectively. We train the model with 400k steps in
ShanghaiTech and 1000k steps in Places2. Besides, $E_{s}$ is trained
with 1/3 of the total steps, and then be fixed. We decay the learning
rate with $0.75$ for each 100k steps. For the structure information,
line segments are extracted by HAWP, and Canny edges are got with
$\sigma=2$. Our model is trained in Pytorch v1.3.1, and costs about
2 days training in ShanghaiTech and about 5 days in Places2 with a
single NVIDIA(R) Tesla(R) V100 16GB GPU.

\noindent \textbf{Comparison Methods.} We compare the proposed MST
with some state-of-the-art methods, which include Gated Convolution
(GC)~\cite{yu2019free}, Edge Connect (EC)~\cite{Nazeri_2019_ICCV},
Recurrent Feature Reasoning (RFR)~\cite{li2020recurrent}, and Mutual
Encoder-Decoder with Feature Equalizations (MED)~\cite{Liu2019MEDFE}.
These methods are all retrained with similar settings and costs compared
with ours. 

\noindent \textbf{Settings of Masks.} 
To handle the real-world image inpainting and editing tasks, such
as object removal, the random irregular mask generation in~\cite{yu2019free,xiong2019foreground}
is adopted in this paper. Besides, as discussed in~\cite{zeng2020high},
real-world inpainting tasks usually remove regions with typical objects
or scenes segments. So we collect 91707 diverse semantic segmentation
masks with various objects and scenes with the coverage rates in $[5\%,40\%]$
from the COCO dataset~\cite{lin2014microsoft}. Overall, the final
mask will be chosen from irregular masks and COCO segment masks randomly
with $50\%$ in both training and test set. To be fair,
all comparison methods are retrained with the same masking strategy.

\subsection{Image Inpainting Results}

For fair comparisons in image inpainting, we do not leak any line
segments of the uncorrupted images for the image inpainting task and
related discussions. The wireframe parser LSM-HAWP is trained in ShanghaiTech,
and it predicts the line segments for the other two datasets. Besides,
results from the object removal with $m=0$ in Eq.~(\ref{2})
are offered for reference only in this section.

\noindent \textbf{Quantitative Comparisons.} In this section, we evaluate
results with PSNR, SSIM~\cite{wang2004image}, and FID~\cite{heusel2017gans}. As discussed in~\cite{zhang2018unreasonable}, we find that $l_{2}$ based metrics such as PSNR and SSIM often contradict human judgment. For example, some meaningless blurring areas will
cause large perceptual deviations but small $l_{2}$ loss~\cite{zhang2018unreasonable}.
Therefore, we pay more attention to the perceptual metric FID in quantitative
comparisons. The results on ShanghaiTech, P2M, and York Urban are
shown in Tab.~\ref{t2}. From the quantitative results, our method outperforms other approaches in PSNR and FID. Especially for the FID, which accords with the human perception, our method
achieves considerable advantages. Besides, our method
enjoys good generalization, as it also works properly in P2M, York Urban, and even P2C.

\begin{figure}
\begin{centering}
\includegraphics[width=0.95\linewidth]{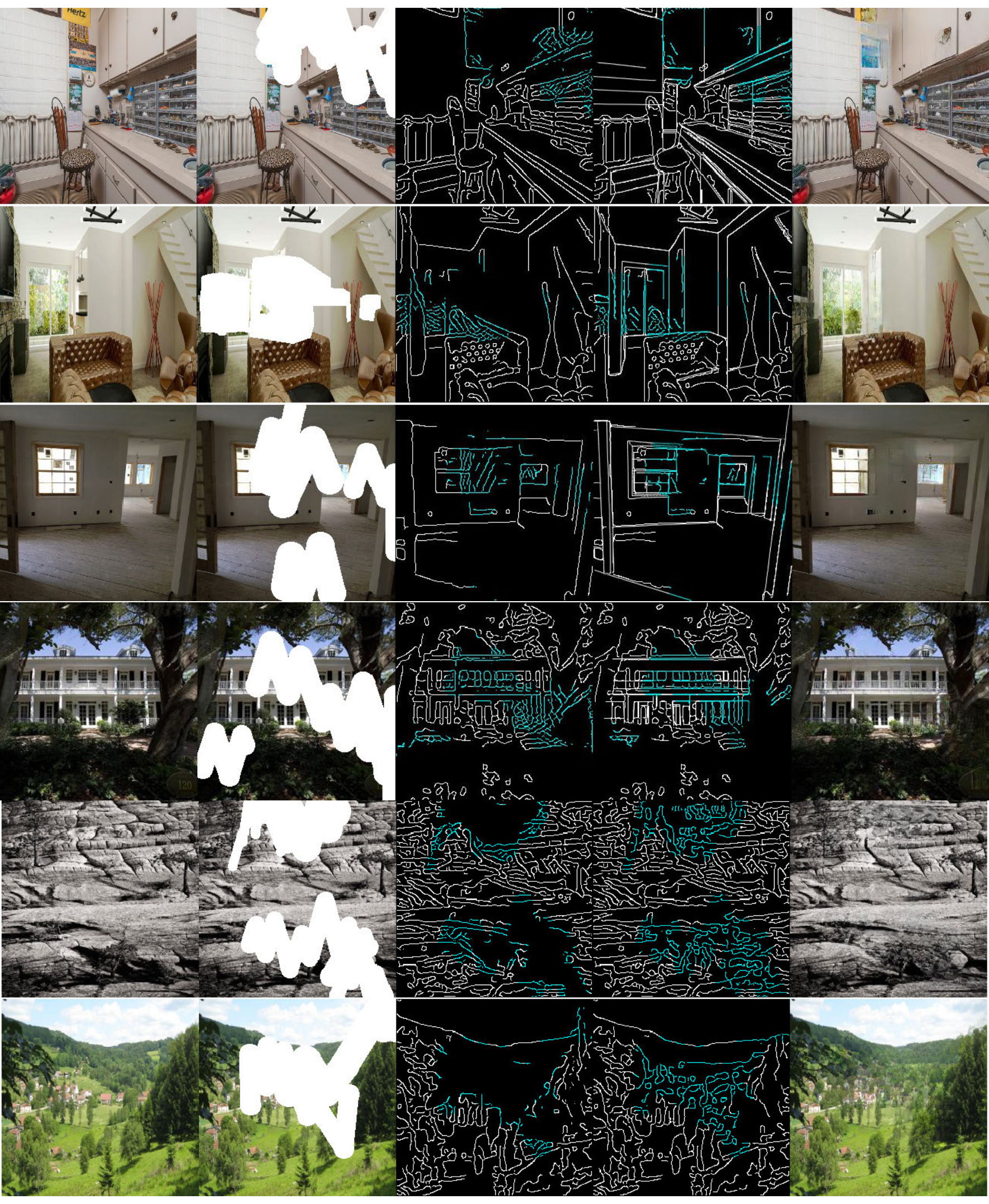} 
\par\end{centering}
\vspace{-0.1in}
 \caption{(From left to right) Original image, masked image, edges $\mathbf{O}_{e}$
generated from EC~\cite{Nazeri_2019_ICCV}, edges and lines $\mathrm{clip}(\mathbf{O}_{l}+\mathbf{O}_{e})$ generated from our model, our inpainted results. Blue lines indicate
the generated lines or edges.}
\label{f6} \vspace{-0.1in}
\end{figure}

\noindent \textbf{Qualitative Comparisons.} Qualitative results among
our method and other state-of-the-art methods are shown in Fig.~\ref{f3}.
Compared with other methods, our approach achieves more semantically coherent
results. From the enlarged regions, our schema significantly outperforms other approaches in preserving the perspectivity and the structures of man-made scenes.
Moreover, we show the edge results of EC~\cite{Nazeri_2019_ICCV} and partial sketch tensor space (edges and lines) results of our method in ShanghaiTech, P2M, and P2C in Fig.~\ref{f6}. 
Results in Fig.~\ref{f6} demonstrate that line segments can supply compositional information to avoid indecisive and meaningless generations in edges.
Furthermore, line segments can also provide clear and specific structures
in masked regions to reconstruct more definitive results in man-made scenes.
Besides, for the natural scene images in P2C without any lines, our method can still outperform EC with  reasonable generated edges for the masked regions. As discussed in Sec.~\ref{sec:PDS}, the proposed PDS works properly to handle the sparse generative problem.

\noindent \textbf{Results of Natural Scenes.} In the last three rows of
Tab.~\ref{t2}, we present the results of ours method on the comprehensive
Places2 dataset (P2C) to confirm the generalization. They are compared
with GC and EC, which have achieved fine scores in man-made Places2
(P2M). Note that all metrics are improved in P2C compared with ones
in P2M of Tab.~\ref{t2}, which demonstrates man-made scenes are
more difficult to tackle. Our methods still get the best results among
all competitors, and the object removal results achieve superior performance.
Besides, the last two rows of Fig.~\ref{f6} show that our method can generate
 reliable edge sketches even without lines in natural scenes.
These phenomenons are largely due to two reasons: 1) There is still a considerable
quantity of line segments in the comprehensive scenes, and these instances
are usually more difficult than others. 2) The proposed partially gated convolutions, efficient attention, and PDS blocks can work properly for various scenes.

\begin{table}
\centering{}{\small{}{}}%
\begin{tabular}{c|ccc}
\hline 
 & \multicolumn{1}{l}{{\small{}{}GC}} & {\small{}{}EC }  & {\small{}{}Ours }\tabularnewline
\hline 
\hline 
{\small{}{}S.-T. }  & {\small{}{}3.00 }  & {\small{}{}7.67 }  & \textbf{\small{}{}33.67}{\small{}{} }\tabularnewline
{\small{}{}P2M }  & {\small{}{}6.67 }  & {\small{}{}8.67 }  & \textbf{\small{}{}32.33}\tabularnewline
\hline 
\end{tabular}
\vspace{-0.05in}
{\small{}{}\caption{Average user scores of ShanghaiTech (S.-T.) and man-made Places2
(P2M).}
\label{table_human_judge} \vspace{-0.2in}
 }{\small\par}
\end{table}

\noindent \textbf{Human Judgements.} For more comprehensive comparisons,
50 inpainted images from GC~\cite{yu2019free}, EC~\cite{Nazeri_2019_ICCV},
and ours are chosen from ShanghaiTech and P2M randomly. And these
samples are compared by 3 uncorrelated volunteers. Particularly, volunteers need to choose the
best one from the mixed pool of the inpainted images with different
methods, and give one score to the respective approach. If all methods
work roughly the same for a certain sample, it will be ignored. The
average scores are shown in Tab.~\ref{table_human_judge}. Ours method
is significantly better than other competitors.

\subsection{Other Applications and Ablation Study}

\begin{figure}
\begin{centering}
\includegraphics[width=0.75\linewidth]{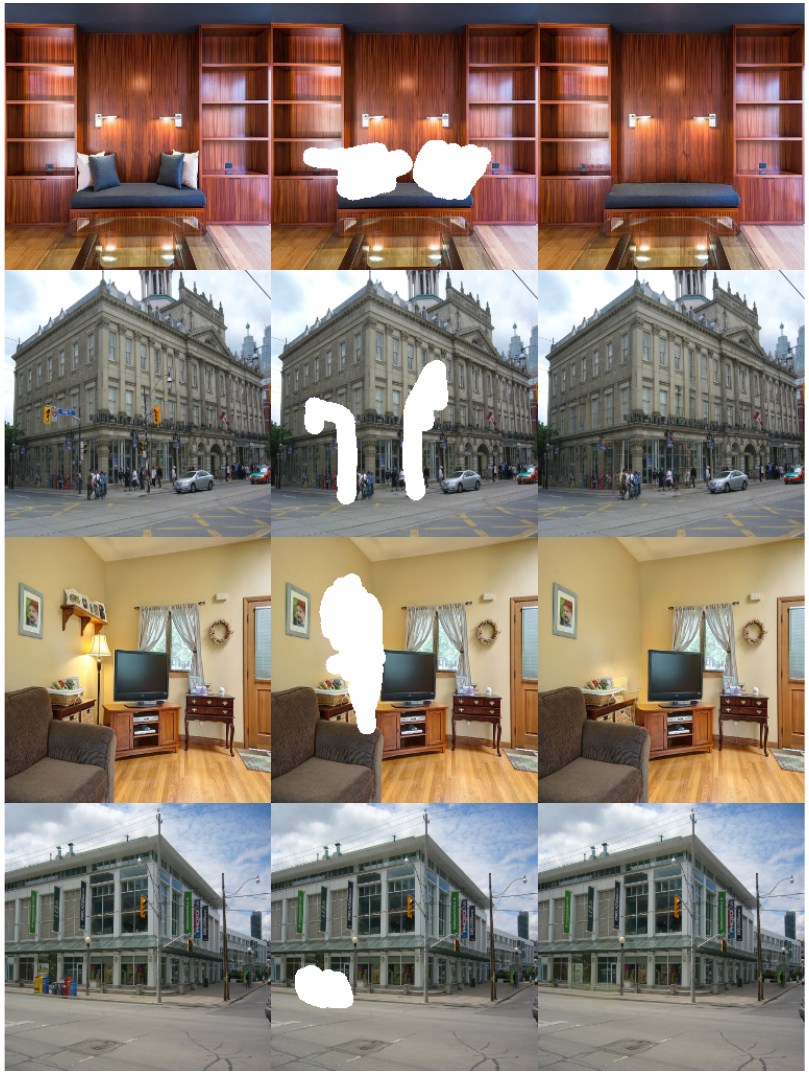} 
\par\end{centering}
\vspace{-0.1in}
 \caption{Object removal examples. From left to right: origin image, masked
image, and our inpainted image.}
\vspace{-0.15in}
 \label{f7} 
\end{figure}

\noindent \textbf{Object Removal}. From the last column in Fig.~\ref{f3},
we show the inpainting results of our MST working in the post-process of
LSM with $m=0$ for the object removal. These refined images with more
reasonable structures indicate that the our model can strengthen
the capability of image reconstruction without redundant residues.
From Fig.~\ref{f7}, some object removal cases are shown to confirm
the practicability of the proposed method. Numerical results are presented
in the last columns of Tab.~\ref{t2}. So, our method can achieve
further improvements in all metrics compared with the vanilla image
inpainting schema for the object removal, and it can correctly mask undesired lines. 
Therefore, the post-process of LSM can significantly improve the performance in the object removal task with the separability of line segments .

\begin{table}
\begin{centering}
\renewcommand\tabcolsep{3.5pt}
{\small{}{}{}}%
\begin{tabular}{c}
{\small{}{}{}\hspace{-0.17in}}
\begin{tabular}{c|ccc|ccc}
\hline 
 & \multicolumn{3}{c|}{{\small{}{}{}unmasked testset}} & \multicolumn{3}{c}{{\small{}{}{}masked testset}}\tabularnewline
\hline 
{\small{}{}{}Threshold}  & {\small{}{}{}5}  & {\small{}{}{}10}  & {\small{}{}{}15}  & {\small{}{}{}5}  & {\small{}{}{}10}  & {\small{}{}{}15}\tabularnewline
\hline 
\hline 
{\small{}{}{}HAWP}  & {\small{}{}{}62.16}  & {\small{}{}{}65.94}  & {\small{}{}{}67.64}  & {\small{}{}{}35.39}  & {\small{}{}{}38.47}  & {\small{}{}{}40.15}\tabularnewline
{\small{}{}{}LSM-HAWP}  & \textbf{\small{}{}{}63.20}{\small{}{} }  & \textbf{\small{}{}{}67.06}{\small{}{} }  & \textbf{\small{}{}{}68.70}{\small{}{} }  & \textbf{\small{}{}{}48.93}{\small{}{} }  & \textbf{\small{}{}{}53.30}{\small{}{} }  & \textbf{\small{}{}{}55.39}\tabularnewline
\hline 
\end{tabular}\tabularnewline
\end{tabular}
\par\end{centering}
\vspace{-0.1in}
 \caption{Structural average precision scores~\cite{zhou2019end} with different
thresholds of parsers trained with (LSM-HAWP) and without (HAWP) the mask augmentation
on ShanghaiTech.}
\vspace{-0.1in}
 \label{t1} 
\end{table}

\begin{table}
\begin{centering}
{\small{}{}{}}
\begin{tabular}{c|cccc}
\hline 
 & {\small{}{}{}PSNR$\uparrow$ }  & {\small{}{}{}SSIM$\uparrow$ }  & {\small{}{}{}FID$\downarrow$ }  & {\small{}{}{}Param }\tabularnewline
\hline 
\hline 
{\small{}{}{}w/o GC }  & {\small{}{}{}26.51 }  & {\small{}{}{}0.871 }  & {\small{}{}{}24.84 }  & {\small{}{}{}10.8M }\tabularnewline
\hline 
{\small{}{}{}Partially GC }  & \textbf{\small{}{}{}26.78}{\small{}{}{} }  & \textbf{\small{}{}{}0.874}{\small{}{}{} }  & {\small{}{}{}22.68 }  & {\small{}{}{}12.1M }\tabularnewline
\hline 
{\small{}{}{}All GC }  & {\small{}{}{}26.71 }  & \textbf{\small{}{}{}0.874}{\small{}{}{} }  & \textbf{\small{}{}{}22.62}{\small{}{}{} }  & {\small{}{}{}25.1M }\tabularnewline
\hline 
\end{tabular}
\par\end{centering}
\vspace{-0.1in}
 \caption{Ablation study of the gated convolution of our model in ShanghaiTech,
Param indicates the parameter scale of the compared inpainting model.}
\label{t4} \vspace{-0.1in}
\end{table}


\begin{table}
\begin{centering}
{\small{}{}{}}%
\begin{tabular}{c}
{\small{}{}{}\hspace{-0.15in}}
\begin{tabular}{c|cccc}
\hline 
 & {\small{}{}{}w/o lines}  & {\small{}{}{}w/o EA}  & {\small{}{}{}w/o PDS}  & {\small{}{}{}Ours} \tabularnewline
\hline 
\hline 
{\small{}{}{}PSNR$\uparrow$}  & {\small{}{}{}26.78}  & {\small{}{}{}26.77}  & {\small{}{}{}26.63}  & \textbf{\small{}{}{}26.90}{\small{}{} }\tabularnewline
\hline 
{\small{}{}{}SSIM$\uparrow$}  & {\small{}{}{}0.874}  & {\small{}{}{}0.875}  & {\small{}{}{}0.873}  & \textbf{\small{}{}{}0.876}{\small{}{} }\tabularnewline
\hline 
{\small{}{}{}FID$\downarrow$}  & {\small{}{}{}22.68}  & {\small{}{}{}21.39}  & {\small{}{}{}21.88}  & \textbf{\small{}{}{}21.16}{\small{}{} }\tabularnewline
\hline 
\end{tabular}\tabularnewline
\end{tabular}
\par\end{centering}
\vspace{-0.1in}
 {\footnotesize{}{}{}\caption{Quantitative ablation studies in ShanghaiTech.}
\label{t5} }{\footnotesize\par}

\vspace{-0.15in}
\end{table}

\noindent \textbf{Masked Wireframe Detection.}
\label{sec:Masked Wireframe Detection} 
As discussed in Sec.~\ref{sec:LSM}, we retrain the HAWP with mask augmentation to ensure the robustness for the corrupted data as LSM-HAWP in ShanghaiTech dataset~\cite{huang2018learning}. 
To confirm the effect of the augmentation, we further prepare another masked ShanghaiTech
testset with the same images. The results are shown in Tab.~\ref{t1}.
The structural average precision (sAP) metric is proposed in~\cite{zhou2019end},
which is defined as the area under the PR curve of detected lines with different thresholds. From Tab.~\ref{t1}, LSM-HAWP performs better than vanilla HAWP significantly for the masked testset. Moreover, HAWP can also gain improvements in the uncorropted testset with more than 1\% in
sAP, which indicates the great generality of the mask augmentation
for the wireframe detection task.


\begin{figure}
\begin{centering}
\includegraphics[width=0.95\linewidth]{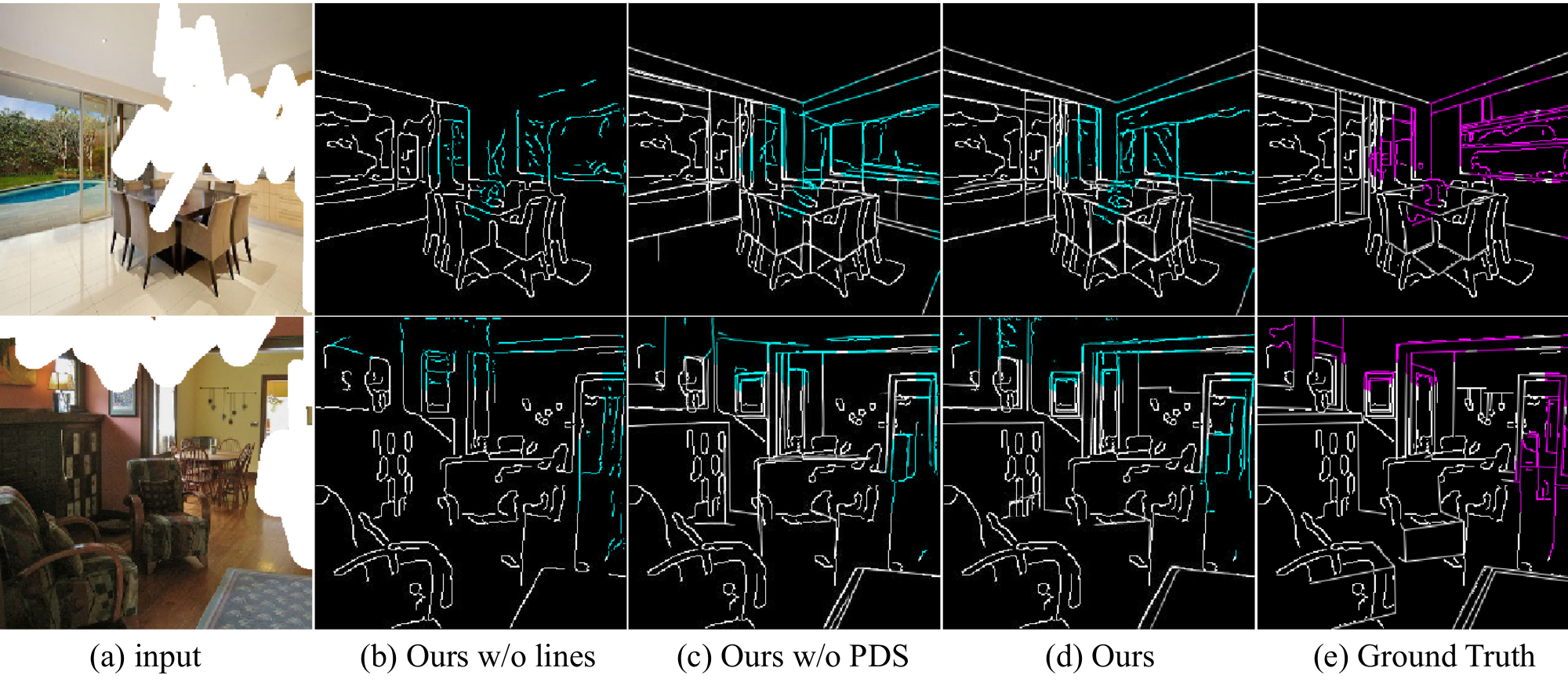} 
\par\end{centering}
\vspace{-0.1in}
 \caption{Ablation studies on line segments and PDS blocks in ShanghaiTech.
Blue and red lines indicate recovered and ground truth in masked regions,
respectively.}
\label{f5} \vspace{-0.15in}
\end{figure}

\noindent \textbf{Simplification of Gated Convolutions.} 
The related ablation study of GC are shown in Tab.~\ref{t4}. 
Only replacing vanilla convolutions for input and output stages of the model with partially GC gains satisfactory improvements. But replacing all convolution layers with GC fails to
achieve a further large advance, while the parameters are doubled\footnote{The visualization results of GC are shown in the supplementary.}.


\noindent \textbf{Contributions of Lines, PDS, and EA.} We explored
the effects of lines, Pyramid Decomposing Separable (PDS) blocks and
the Efficient Attention (EA) module in Tab.~\ref{t5} in ShanghaiTech. Specifically,
line segments, PDS blocks and the EA module are removed respectively,
while other settings unchanged. 
As shown in Fig.~\ref{f5}, our method without line segments causes
severe structural loss, and the one without PDS suffers from contradictory
perspectivity. From Tab.~\ref{t5}, we can see all line segments,
PDS blocks and the EA module can improve the performance in image
inpainting.

\section{Conclutions}

This paper studies an encoder-decoder MST-net for inpainting. It learns a sketch tensor space
restoring edges, lines, and junctions.
Specifically, the encoder extracts and refines line and edge structures into a sketch tensor space. The decoder recover the image from it.  
Moreover, the LSM algorithm is proposed to update the line extractor HAWP to fit the inpainting and object removal tasks.
Several effective network modules are proposed to improve the MST for tough man-made scenes inpainting. 
Extensive experiments validate the efficacy of our MST-net for inpainting.

\noindent {\small \textbf{Acknowledgment} Yanwei Fu is the corresponding authour. This project is supported by Huawei, and released on \url{https://ewrfcas.github.io/MST_inpainting}. }

{\small
\bibliographystyle{ieee_fullname}
\bibliography{egbib}
}

\section{Appendix}

\subsection{Network Architectures}

The network architectures  are illustrated as follows. 

\noindent\textbf{Partially Gated Convolution (GC) Block.} Both encoder and decoder contain two groups of partially GC blocks for the input (3 blocks) and output (2 blocks) features, and the GC is consist of GateConv2D~\cite{yu2019free} $\rightarrow$ InstanceNorm $\rightarrow$ ReLU. We also add the spectral normalization~\cite{miyato2018spectral} to the GC used in the generator.

\noindent\textbf{Dilated Residual Block.}
The dilated residual blocks are the same as the ones used in EdgeConnect~\cite{Nazeri_2019_ICCV} with Conv2D $\rightarrow$ InstanceNorm $\rightarrow$ ReLU $\rightarrow$ Conv2D $\rightarrow$ InstanceNorm, where the first convolution is based on $\mathrm{dilation}=2$.

\noindent\textbf{Efficient Attention.}
The input feature of the Efficient Attention is $256$ channels, and we use the multi-head attention with $n_{head}=4$. So, the dimension $d'$ of each group is $256/4=64$.

\subsection{Details of the Experiments}

\subsubsection{Preprocessing}

For our method, the input images are resized to 256$\times$256 at first. Then they are normalized to $[-1, 1]$, and values of the masked regions are set to 1. For the line detection, we set the thresholds of LSM-HAWP for unmasked and masked with (0.95, 0.925) in ShanghaiTech~\cite{huang2018learning}, and (0.85, 0.8) in Places2~\cite{zhou2017places} and York Urban~\cite{denis2008efficient}.

\subsubsection{Places2 Dataset Selection}

\begin{figure}
\begin{centering}
\includegraphics[width=0.95\linewidth]{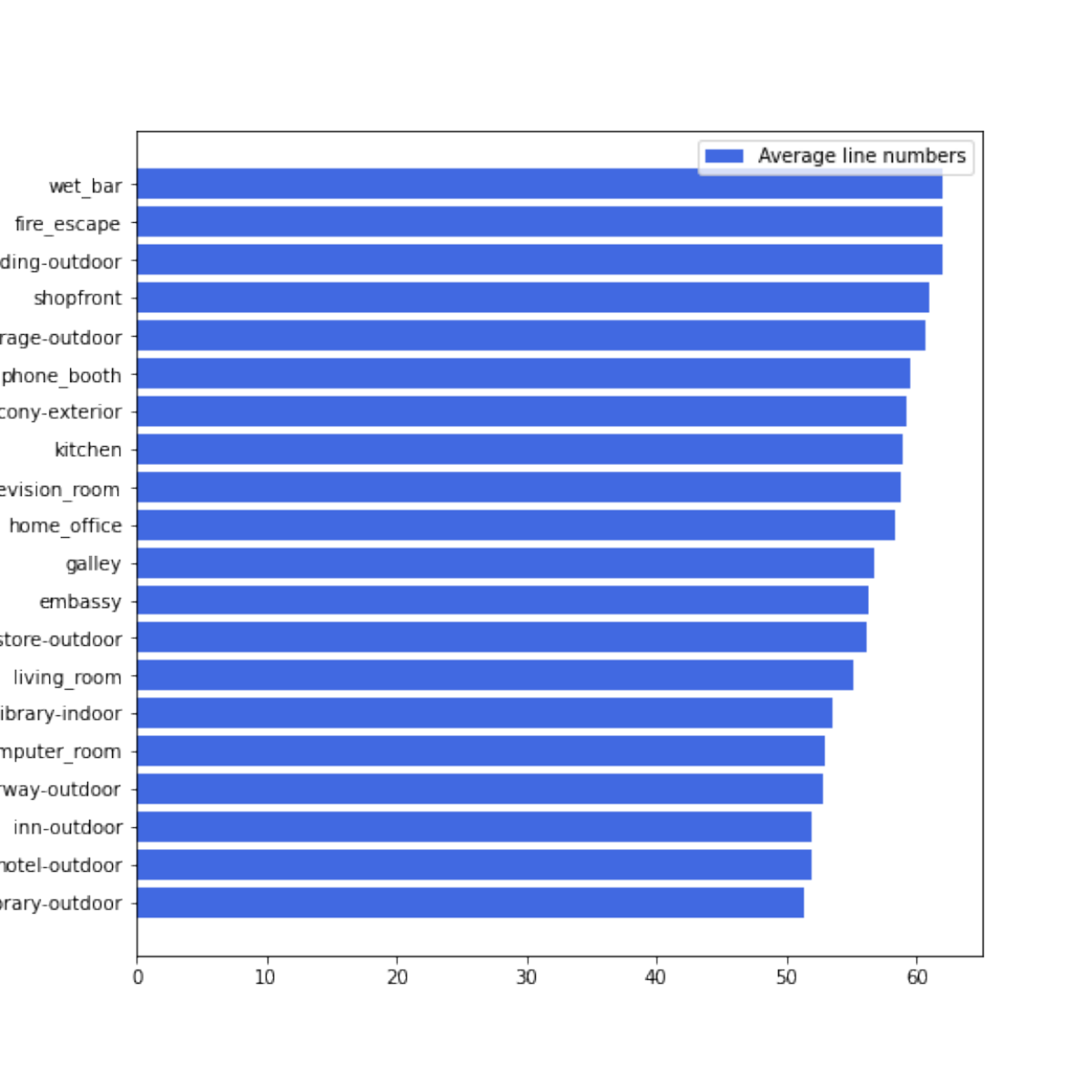} 
\par\end{centering}
\caption{The bar chart of the scenes with top20 average line segment (confidence$\geq0.925$) numbers of Places2.}
\label{fig:places2_line_numbers} 
\end{figure}

\begin{figure}
\begin{centering}
\includegraphics[width=0.9\linewidth]{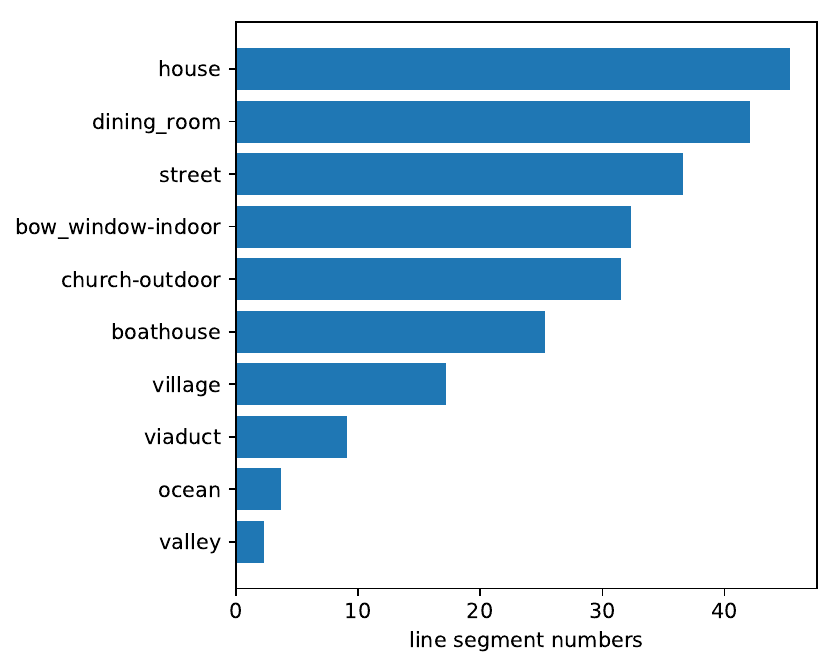} 
\par\end{centering}
\caption{The bar chart of the line segment (confidence$\geq0.925$) numbers of the comprehensive Places2 (P2C).}
\label{fig:P2C_line_segments} 
\end{figure}

Places2 dataset~\cite{zhou2017places} is consisted of 10 million+ images from 365 various scenes. Since the proposed method is devoted to reconstructing the images with man-made structures, we need to select some representative scenes whose images contain enough structure information for man-made Places2 (P2M). Therefore, we leverage the retrained LSM-HAWP to predict the number of extracted line segmentations with scores larger than 0.925 in the validation of Places2. The scenes with top20 average line segment numbers are shown in Fig.~\ref{fig:places2_line_numbers}. For the generalization, we randomly select 10 of them which is consist of `balcony-exterior', `computer-room', `embassy', `galley', `general-store-outdoor', `home-office', `kitchen', `library-outdoor', `parking-garage-outdoor', and `shopfront'. And there are 47949 training images and 1000 test images at all, which include buildings, indoor scenes, and outdoor scenes with various structures.

Besides, we also provide the results of comprehensive Places2 (P2C) in the main paper, which is consist of 10 randomly selected scenes: `valley', `church-outdoor', `village', `house', `dining-room', `street', `ocean', `bow-window-indoor', `boathouse', and `viaduct'. And the related line segments are shown in Fig.~\ref{fig:P2C_line_segments}.

\subsubsection{Implements of Compared Methods}
All of the compared methods are based on the official implements with the new mask strategy mentioned in the paper. The hyper-parameter settings and learning rate settings refer to the official settings too. Besides, other settings are listed as follows. Note that code addresses listed in the footnote are the official implements of other compared methods. 

\noindent\textbf{Gated Convolution (GC)}~\cite{yu2019free}\footnote{\url{https://github.com/JiahuiYu/generative_inpainting}} GC is trained with 1500 epochs (about 468k steps) for ShanghaiTech, and 350 epochs (1049k steps) for Places2 with batch size 16.

\noindent\textbf{Edge Connect (EC)}~\cite{Nazeri_2019_ICCV}\footnote{\url{https://github.com/knazeri/edge-connect}}
EC is trained with 400k (150k for the edge refinement network) steps for ShanghaiTech, and 1000k (300k steps for the edge refinement network) for Places2 with batch size 16.

\noindent\textbf{Recurrent Feature Reasoning (RFR)}~\cite{li2020recurrent}\footnote{\url{https://github.com/jingyuanli001/RFR-Inpainting}}
RFR is trained with 400k (150k for finetune) steps for ShanghaiTech, and 1200k (200k for finetune) for Places2 with batch size 6.

\noindent\textbf{Mutual Encoder Decoder with Feature Equalizations (MED)}~\cite{Liu2019MEDFE}\footnote{\url{https://github.com/KumapowerLIU/Rethinking-Inpainting-MEDFE}}
Note that the official implement of MED only supports training with batch=1 due to its special CSA attention design~\cite{liu2019coherent}. So, the training is very inefficient and it takes us more than three weeks to train the MED on Places2. However, MED still works badly on Places2, which dues to the instable training process with batch=1 in our opinions.
As the result, MED is trained with 200 epochs (1000k steps) for ShanghaiTech, and 60 epochs (about 2877k steps) for Places2 with batch size 1.

\subsection{Supplementary Experimental Results}

\subsubsection{Visualization of Gated Convolutions}

\begin{figure}
\begin{centering}
\includegraphics[width=0.95\linewidth]{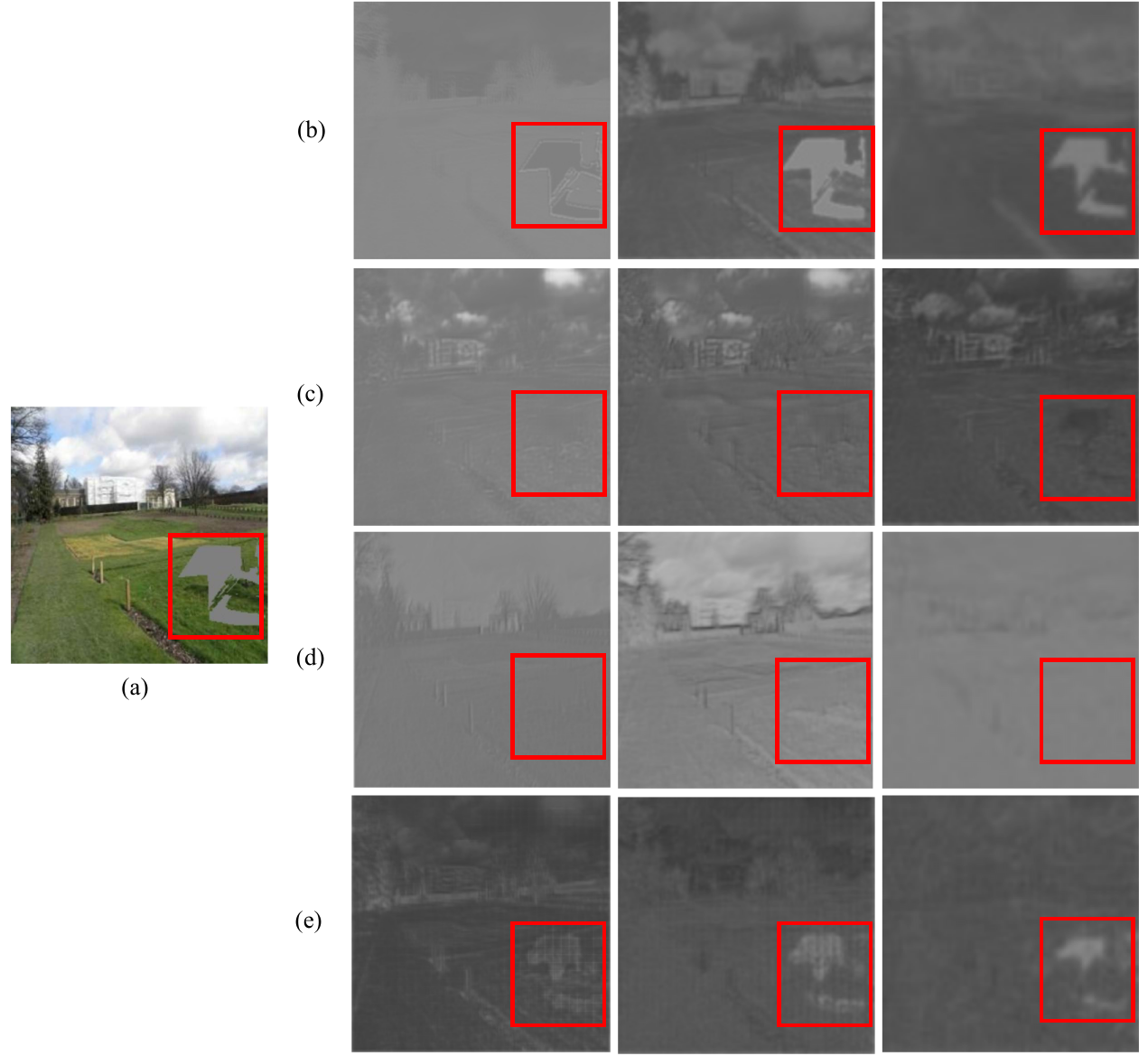} 
\par\end{centering}
\caption{Visualization of sigmoid weight outputs $\sigma(\mathbf{G})$
in GC: (a) is the masked input, and (b), (c), (d), (e) are $\sigma(\mathbf{G})$ with different scales
from the coarse network encoder, coarse network decoder, refinement network encoder, and refinement network decoder respectively.}
\label{fig:gc_vis} 
\end{figure}

We show the visualization results of the origin GC in Fig.~\ref{fig:gc_vis}, which are
collected by averaging the $\sigma(\mathbf{G})$ outputs from different GateConvs.
$\sigma(\mathbf{G})$ for the masked regions are active in Fig.~\ref{fig:gc_vis}(b) and Fig.~\ref{fig:gc_vis}(e),
which are the encoder of the coarse network and the decoder of the refinement network. And they can also be seen as the encoder and decoder of the whole GC pipeline model. 

\subsubsection{Efficiency Comparisons of Efficient Attention}

\begin{table}
\small
\begin{center}
\begin{tabular}{ccc}
\toprule
Attention & Batchsize & Image/sec\tabularnewline
\midrule
CA & 16 & 19.42\tabularnewline
EA & 16 & 36.37\tabularnewline
CA & 32 & 20.70\tabularnewline
EA & 32 & 40.81\tabularnewline
\bottomrule
\end{tabular}
\end{center}
\caption{Comparison of the training speed of our model based on Contextual Attention (CA)~\cite{yu2018generative} and Efficient Attention (EA)~\cite{shen2018efficient} with different batch sizes.}
\label{tab:att} 
\end{table}

In this section, we pay attention to compare the efficiency of Contextual Attention (CA)~\cite{yu2018generative} and Efficient Attention (EA)~\cite{shen2018efficient}. All comparisons are based on our structure refinement network, and only one attention layer is added to the middle of the residual blocks with the resolution 64$\times$64. Then, we train the model with two attention strategies and batch sizes on the ShanghaiTech dataset for one epoch. The results are shown in Tab.~\ref{tab:att}. Since CA can only be implemented without parallel computing for the batch, we just compare the training speed with images per second between CA and EA. From Tab.~\ref{tab:att}, EA can be trained more efficiently than CA. Moreover, benefited from the parallelism, EA enjoys the speedup by enlarging the training batch size.

\subsubsection{Ablation Study with Only Edge and PSS}

\begin{figure}
\begin{centering}
\includegraphics[width=0.95\linewidth]{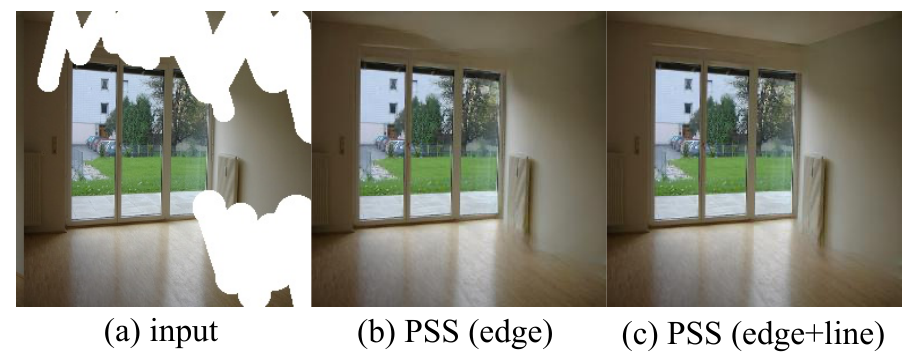} 
\par\end{centering}
\caption{Qualitative results w. and w./o. lines in ShanghaiTech.}
\label{fig:only_edge} 
\end{figure}

\begin{table}
\small
\begin{center}
\begin{tabular}{cccc}
\toprule
 & PSNR$\uparrow$ & SSIM$\uparrow$ & FID$\downarrow$\tabularnewline
\midrule
PSS (edge) & 26.78 & 0.875 & 21.40\tabularnewline
PSS (edge+line) & \textbf{26.90} & \textbf{0.876} & \textbf{21.16}\tabularnewline
\bottomrule
\end{tabular}
\end{center}
\caption{Quantitative results w. and w./o. lines in ShanghaiTech.}
\label{tab:only_edge} 
\end{table}

To further improve the effectiveness of lines extracted from the LSM-HAWP. Here we provide the qualitative and quantitative results in Fig.~\ref{fig:only_edge} and Tab.~\ref{tab:only_edge}: only canny edges with multi-scale training of PSS (b), V.S. our full model (c). Some lines on the roof are not recovered in (b) of Fig.~\ref{fig:only_edge}.

\subsubsection{Comparisons with ProFill}

\begin{figure}
\begin{centering}
\includegraphics[width=1.0\linewidth]{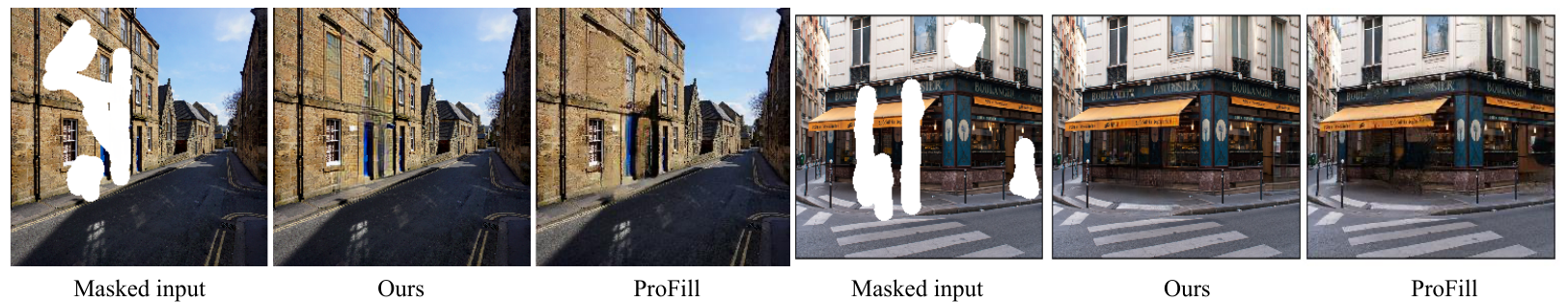} 
\par\end{centering}
\caption{Qualitative results compared with ProFill in flickr.}
\label{fig:vsprofill} 
\end{figure}

ProFill~\cite{zeng2020high} is one of SOTA inpainting methods, but no open source codes/models of ProFill are available now. To ensure the fairness, we compare qualitative results from flickr by the API of ProFill in Fig.~\ref{fig:vsprofill}, which are independent from the training set of both our method and ProFill.

\subsubsection{Additional Visual Results}

In this section we provide more experiment results for image inpainting and object removal on different datasets. Furthermore, we provided more comparisons in P2C compared with EC, which show the superior performance of the proposed sketch tensor space reconstructing.

\subsubsection{Object Removal Video}

We also provide a video that displays the object removal in ShanghaiTech, P2M, and P2C. The video is played in a triple-speed to compress the storage size. This video shows the good generalization and the fast interactive experience of the proposed method in the GPU environment.

\begin{figure*}
\begin{centering}
\includegraphics[width=0.9\linewidth]{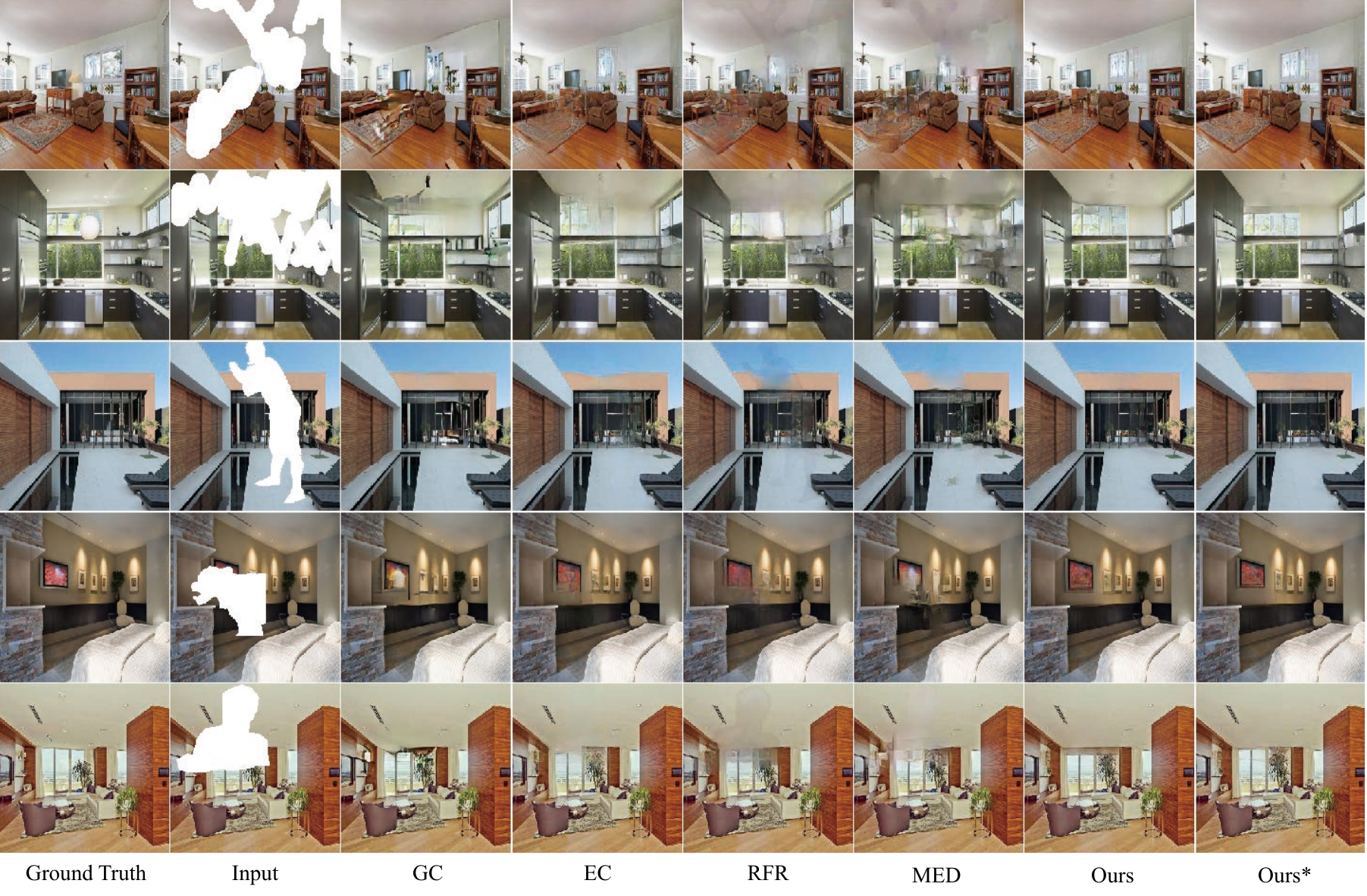} 
\par\end{centering}
\caption{Inpainting comparisons on ShanghaiTech, where * means that our method works in the object removal mode.}
\label{fig:shanghaitech_comparisons} 
\end{figure*}

\begin{figure*}
\begin{centering}
\includegraphics[width=0.9\linewidth]{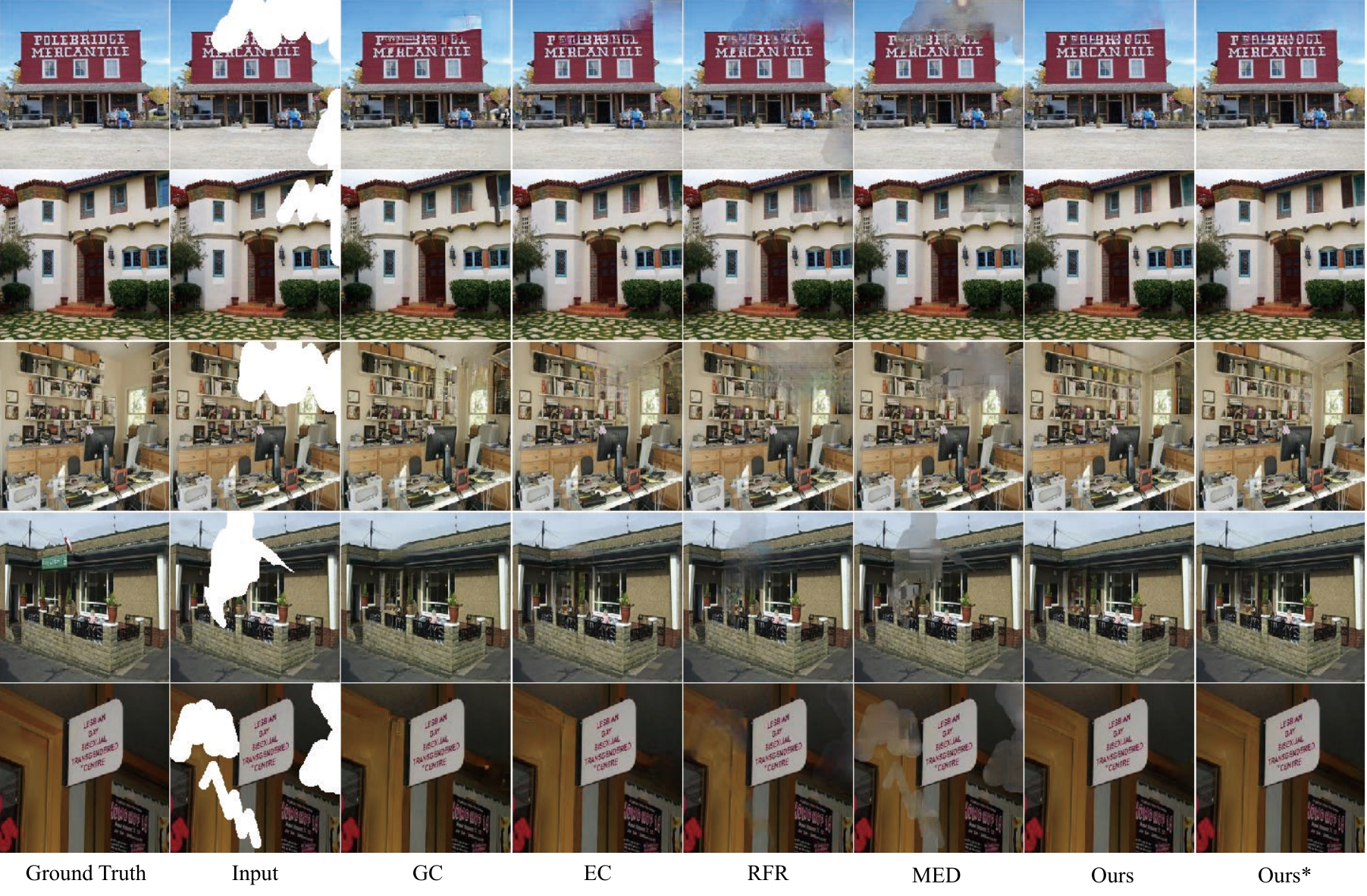} 
\par\end{centering}
\caption{Inpainting comparisons on P2M, where * means that our method works in the object removal mode.}
\label{fig:places2_comparisons} 
\end{figure*}

\begin{figure*}
\begin{centering}
\includegraphics[width=0.925\linewidth]{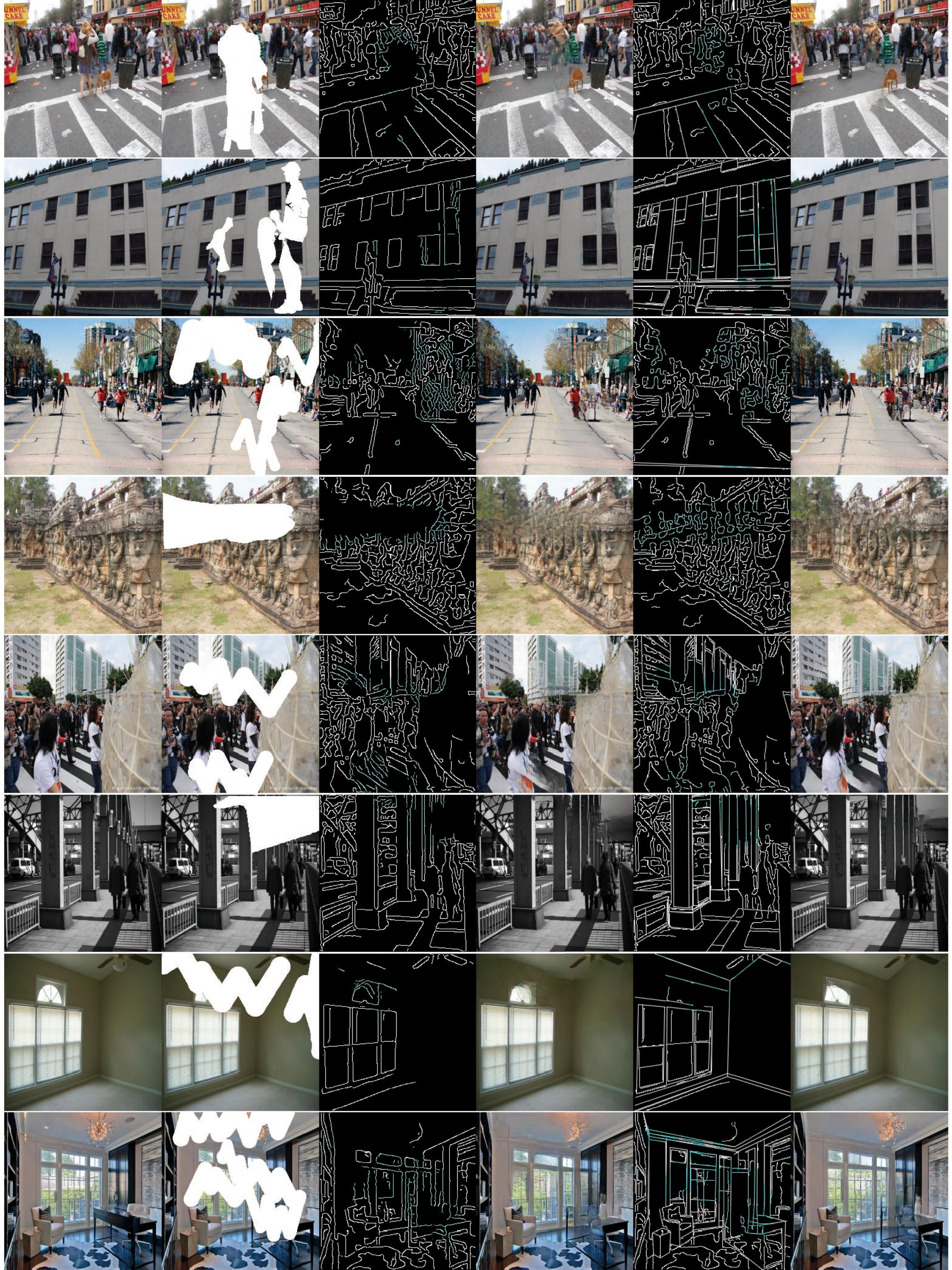} 
\par\end{centering}
\caption{Inpainting results in P2C compared with EC (part1). From left to right: origin image, input masked image, refined edges from EC, inpainted results of EC, our refined sketch tensors, our inpainted results (generated lines and edges are blue).}
\label{fig:p2c_compare1} 
\end{figure*}

\begin{figure*}
\begin{centering}
\includegraphics[width=0.925\linewidth]{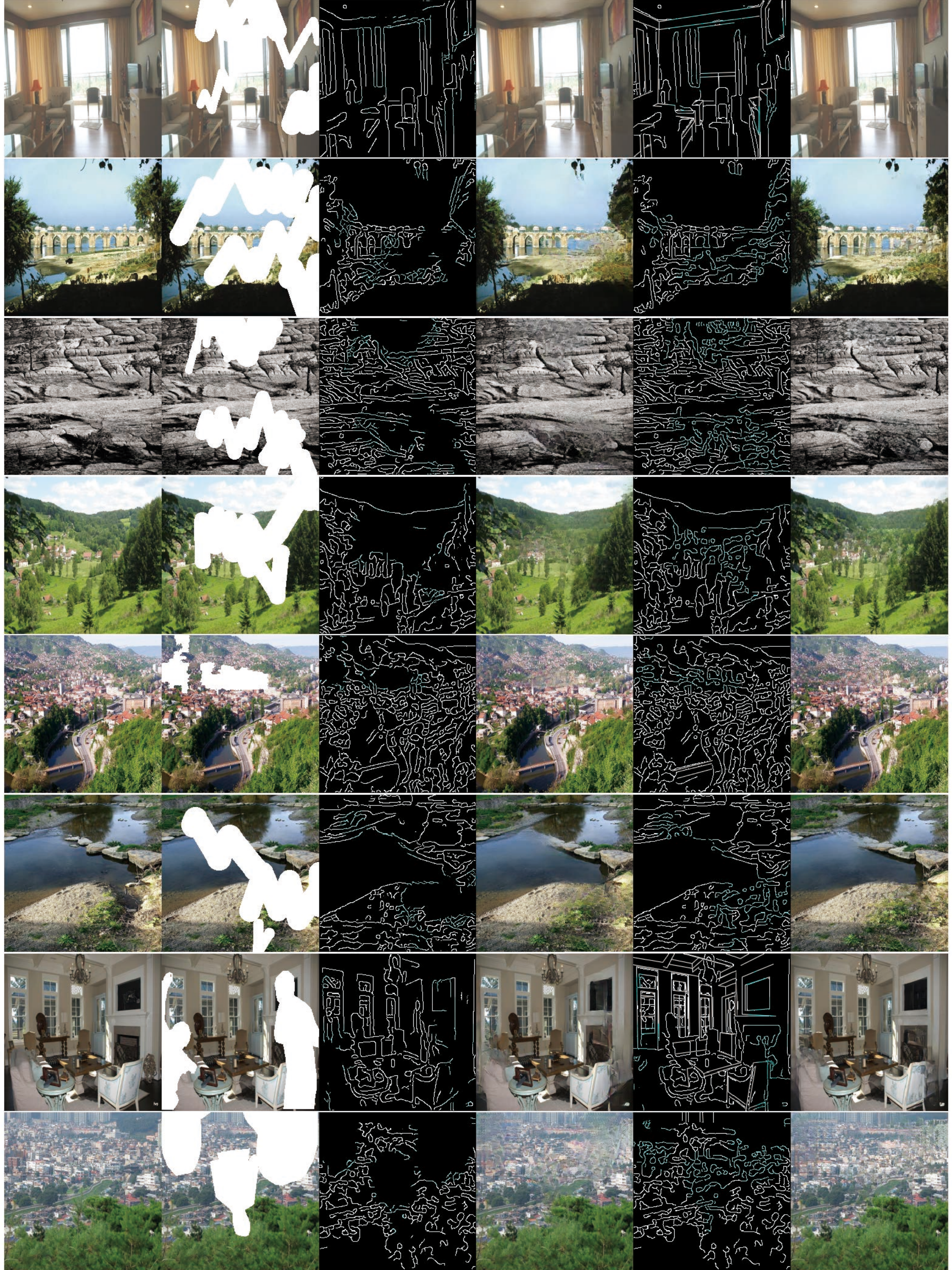} 
\par\end{centering}
\caption{Inpainting results in P2C compared with EC (part2). From left to right: origin image, input masked image, refined edges from EC, inpainted results of EC, our refined sketch tensors, our inpainted results (generated lines and edges are blue).}
\label{fig:p2c_compare2} 
\end{figure*}

\begin{figure*}
\begin{centering}
\includegraphics[width=1\linewidth]{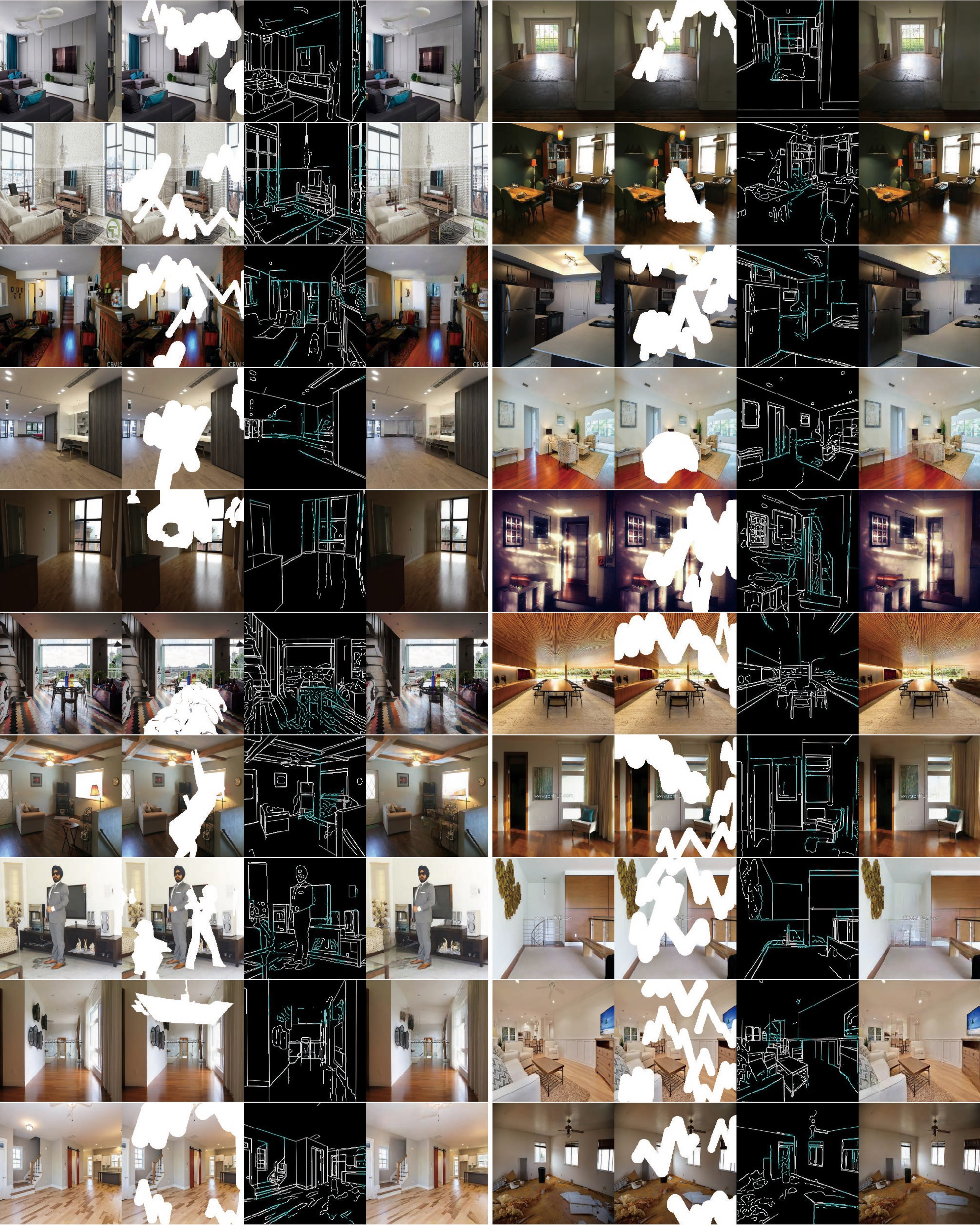} 
\par\end{centering}
\caption{Inpainting results of our method on ShanghaiTech. For each of the two columns, from left to right: origin image, input masked image, refined structures (generated lines and edges are blue), inpainted results.}
\label{fig:shanghaitech_inpainting} 
\end{figure*}

\begin{figure*}
\begin{centering}
\includegraphics[width=1\linewidth]{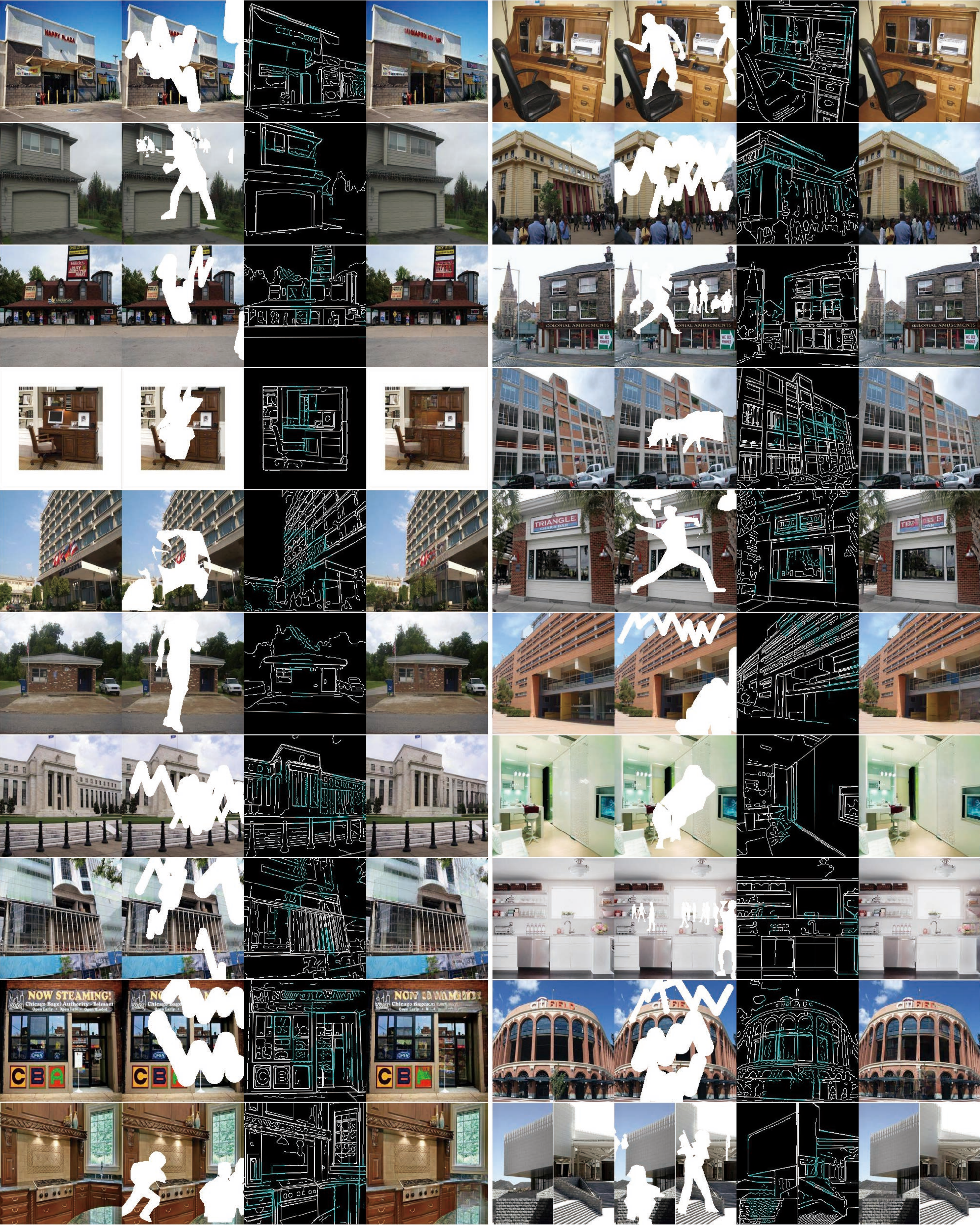} 
\par\end{centering}
\caption{Inpainting results of our method on P2M. For each of the two columns, from left to right: origin image, input masked image, refined structures (generated lines and edges are blue), inpainted results.}
\label{fig:places2_inpainting} 
\end{figure*}

\begin{figure*}
\begin{centering}
\includegraphics[width=1\linewidth]{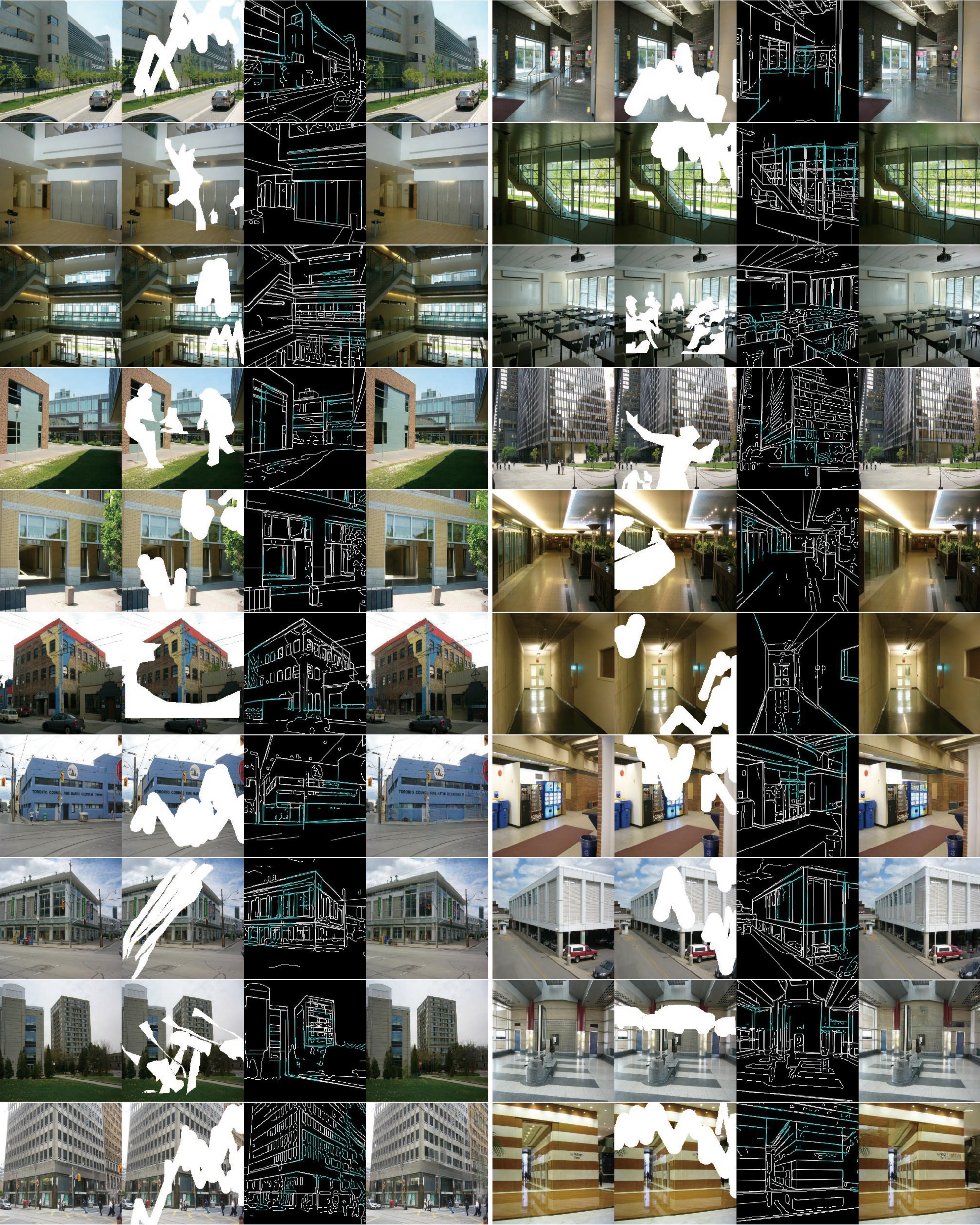} 
\par\end{centering}
\caption{Inpainting results of our method on York Urban. For each of the two columns, from left to right: origin image, input masked image, refined structures (generated lines and edges are blue), inpainted result.}
\label{fig:york_inpainting} 
\end{figure*}

\begin{figure*}
\begin{centering}
\includegraphics[width=1\linewidth]{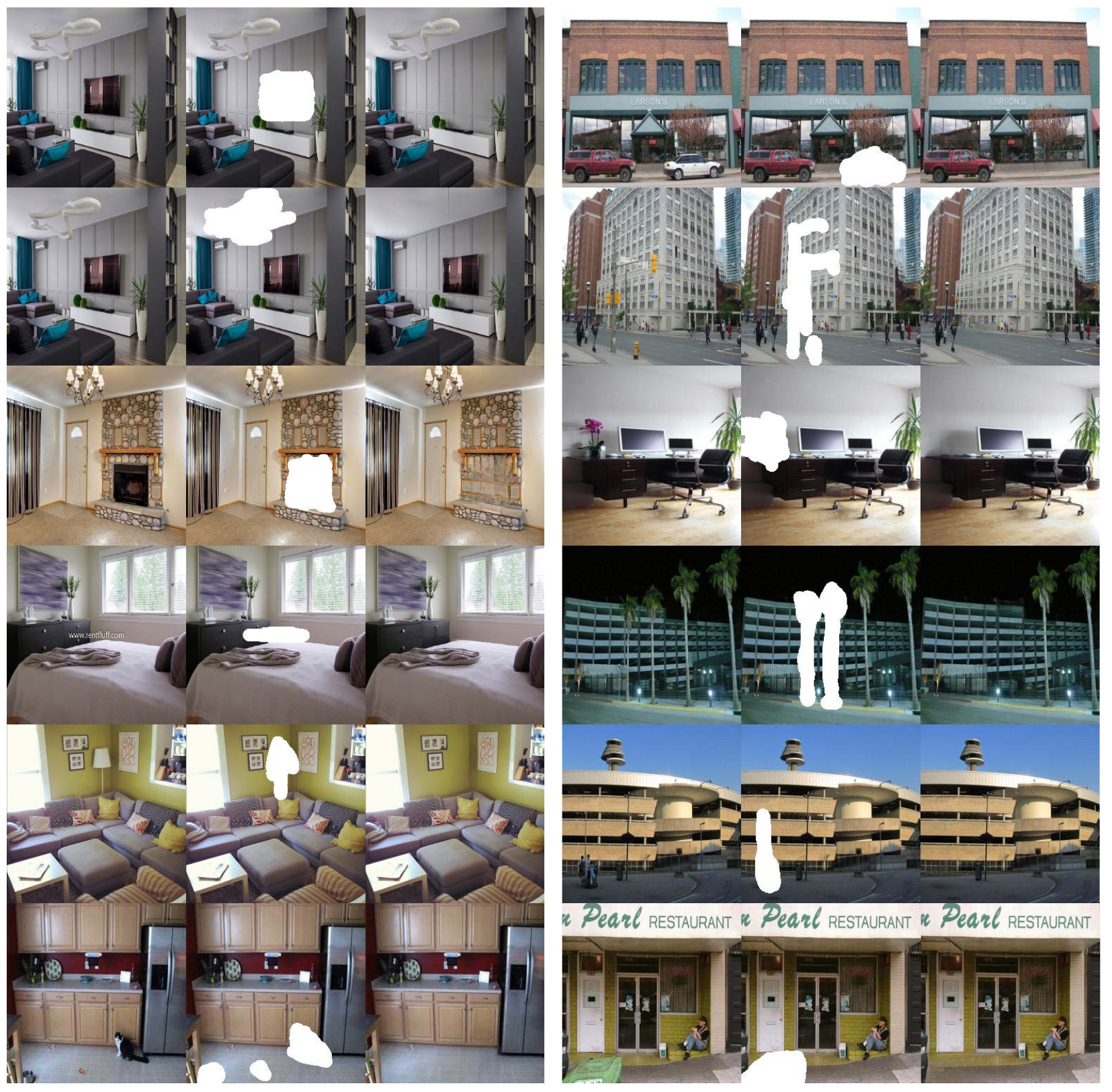} 
\par\end{centering}
\caption{Object removal results from ShanghaiTech, Places2, and York Urban. For each of the two columns, from left to right: origin image, input masked image, generated image.}
\label{fig:object_removal} 
\end{figure*}

\end{document}